%% file: main.tex
\documentclass[accepted]{uai2025} % after acceptance, for a revised version; 
\usepackage[american]{babel}

\usepackage{natbib} % has a nice set of citation styles and commands
\usepackage[utf8]{inputenc}
\usepackage[T1]{fontenc}
\usepackage{amsfonts}
\usepackage{nicefrac} 
\usepackage{microtype}
\usepackage{tabularx}
\usepackage{graphicx}
\usepackage{amsmath}
\usepackage{amsthm}
\usepackage{amssymb}
\usepackage{mathtools}
\usepackage{float}
\usepackage{booktabs}
\usepackage{bbm}
\usepackage{bm}
\usepackage{multirow}
\usepackage[dvipsnames, table]{xcolor}
\usepackage{subcaption}
\usepackage{enumitem}
\usepackage{url}
\include{text/def_math_cmds}

\hypersetup{colorlinks=true, allcolors=blue}

\newcommand{\ie}{\emph{i.e.}}
\newcommand{\eg}{\emph{e.g.}}

\newtheorem{theorem}{Theorem}[section]

\newtheorem{definition}[theorem]{Definition}

\title{CP$^2$: Leveraging Geometry for Conformal Prediction via Canonicalization}

\author[1]{Putri A. van der Linden\thanks{Corresponding author: \href{mailto:p.a.vanderlinden@uva.nl}{p.a.vanderlinden@uva.nl}}$^{,}$}
\author[1,2]{Alexander Timans}
\author[1]{Erik J. Bekkers}

\affil[1]{%
    Amsterdam Machine Learning Lab, University of Amsterdam
} \affil[2]{%
    UvA-Bosch Delta Lab, University of Amsterdam
}

\begin{document}
\maketitle

%%%%%%%%%%%%%%%%%%%%%%%%%%%%%%%%
% TEXT
%%%%%%%%%%%%%%%%%%%%%%%%%%%%%%%%

\input{text/abstract}

\section{Introduction}
\label{sec:intro}
\input{text/intro_v2}

\section{Background}
\label{sec:background}
\input{text/background}

\section{Geometric Information for Conformal Prediction}
\label{sec:method}
\input{text/method}

\section{Experiments}
\label{sec:exp}
\input{text/exp}

\section{Related Work}
\label{sec:related-work}
\input{text/related_work}

\section{Discussion}
\label{sec:discussion}
\input{text/discussion}

% Acknowledgments, Societal Impact
\input{text/ack_impact}

% \clearpage
% References
\bibliography{main}

%%%%%%%%%%%%%%%%%%%%%%%%%%%%%%%%
% APPENDIX
%%%%%%%%%%%%%%%%%%%%%%%%%%%%%%%%

\clearpage
\appendix
\input{text/appendix}

\end{document}

%% file: text/def_math_cmds.tex
\usepackage{amsmath,amsfonts,bm}
\usepackage{upgreek}

\def\eqref#1{equation~\ref{#1}}
\def\1{\bm{1}}

\def\rvx{{\mathbf{x}}}
\def\rvy{{\mathbf{y}}}

\def\vtheta{{\bm{\theta}}}

\def\vx{{\bm{x}}}
\def\vy{{\bm{y}}}

\DeclareMathAlphabet{\mathsfit}{\encodingdefault}{\sfdefault}{m}{sl}
\SetMathAlphabet{\mathsfit}{bold}{\encodingdefault}{\sfdefault}{bx}{n}

\def\gD{{\mathcal{D}}}

\def\gX{{\mathcal{X}}}
\def\gY{{\mathcal{Y}}}

%% file: text/abstract.tex
\begin{abstract}
    We study the problem of \emph{conformal prediction} (CP) under geometric data shifts, where data samples are susceptible to transformations such as rotations or flips. While CP endows prediction models with \emph{post-hoc} uncertainty quantification and formal coverage guarantees, their practicality breaks under distribution shifts that deteriorate model performance. To address this issue, we propose integrating geometric information---such as geometric pose---into the conformal procedure to reinstate its guarantees and ensure robustness under geometric shifts. In particular, we explore recent advancements on pose \emph{canonicalization} as a suitable information extractor for this purpose. Evaluating the combined approach across discrete and continuous shifts and against equivariant and augmentation-based baselines, we find that integrating geometric information with CP yields a principled way to address geometric shifts while maintaining broad applicability to black-box predictors.
\end{abstract}

% \putri{To address this issue, we propose integrating sample-wise geometric information --- such as geometric pose --- into our conformal procedures to reinstate its guarantees and ensure robustness under geometric shifts. In this work we explore the canonicalization principle \citep{kaba23equivariance, mondal2023equivariant} as a light-weight geometry extractor, and show that our proposal preserves CP’s \emph{post-hoc} flexibility while mitigating performance degradation, providing information for conditional coverage, and enabling adaptive weighting for multi-shift scenarios.}

%% file: text/intro_v2.tex
The deployment of machine learning models---including deep neural networks---has become increasingly widespread, yet their application in costly or safety-critical settings remains hindered by two key challenges \citep{makridakis2016forecasting,quinonero2022dataset}. Firstly, many models continue to produce point-wise predictions without uncertainty estimation, inherently limiting the robustness of obtained information for decision-making \citep{begoli2019need, padilla2021uncertain}. Yet even when uncertainty is incorporated, such as through some form of uncertainty scoring or probabilistic modelling \citep{gawlikowski2023survey}, estimates can be misleading or overconfident \citep{kompa2021empirical, xiong2023can}. A popularized uncertainty framework that partially addresses such issues is \emph{conformal prediction} (CP), which extends point-wise predictions to prediction set or interval estimation \citep{vovk2005algorithmic, angelopoulos2023conformal}. Importantly, a notion of reliability is obtained via a probabilistic coverage guarantee for new, unseen test samples (see \autoref{subsec:background-cp}). Unlike traditional prediction set methods \citep{khosravi2011comprehensive}, CP is fully data-driven, distribution-free, and compatible with `black-box' models.

Secondly, it is well-known that distribution shifts at test time can severely degrade model performance \citep{koh2021wilds, ovadia2019can}. Among types of shifts, \emph{geometric} data shifts---where test samples undergo geometric transformations such as rotations or flips---pose a significant challenge, in particular for pretrained models lacking integrated equivariance or invariance properties \citep{bronstein2021geometric}. As such symmetry-awareness can be sometimes challenging to scale and is thus overlooked \citep{brehmer2024does}, large models trained on vast datasets may nonetheless struggle when faced with pose variations, as exemplified in \autoref{tab:segm-robustness} for segmentation under rotations. Other practical failures may include proper recognition for medical images due to scan variations \citep{fu2023guest} or 3D objects due to axis-misaligned point clouds \citep{vadgama2025utilityequivariancesymmetrybreaking}. For conformal prediction, such geometric shifts can violate \emph{exchangeability} assumptions on the data (\hyperref[def:exch]{Def.~\ref{def:exch}}), leading to potentially unreliable or uninformative prediction sets \citep{barber2023conformal}. Unreliable in the sense that statistical coverage guarantees may no longer hold, and uninformative as prediction sets may grow excessively large. 

To address this, we propose robustifying the conformal procedure by incorporating geometric information on occuring shifts, while preserving CP's advantageous flexibility by avoiding to modify the underlying model. This is practically achieved via \emph{canonicalization} \citep{mondal2023equivariant, kaba23equivariance}, a framework that learns to map data into a canonical form, and decouples the geometric task from the underlying predictor. Leveraging this approach, we explore how obtained geometric information can be effectively combined with CP in multiple different ways. In summary, our contributions include:
\begin{itemize}
    \item Introducing a novel geometric perspective on the topic of distribution shifts in conformal prediction, and motivating how geometric information can ensure core conditions of CP such as exchangeability are met (\autoref{sec:method});
    \item Leveraging canonicalization as a suitable geometric information extractor that is both \emph{post-hoc} and light-weight, in line with practical principles underlying CP;
    \item Investigating its integration with CP in several ways, including mitigating performance drops (\autoref{subsec:exp-robust}), as an information tool for conditional coverage (\autoref{subsec:exp-condcover}), and as a weighting mechanism in multi-shift settings (\autoref{subsec:exp-weightcp}).
\end{itemize}

\begin{table}[t]
  \caption{
    Zero-shot Mask-RCNN segmentation performance (mAP) on regular and $C4$-rotated COCO data without and with invariance (via canonicalization \citep{mondal2023equivariant}). Missing symmetry-awareness leads to failed generalization.
  }
  \centering
  \begin{tabularx}{\linewidth}{X|c|c}
    \toprule
     \textbf{Model} & \textbf{mAP} & \textbf{$C4$-mAP} \\
     \midrule
    Mask-RCNN without Invariance  & 47.81 & 12.79 \\
    Mask-RCNN with Invariance & 43.47 & 43.47 \\
    \bottomrule
  \end{tabularx}
  \label{tab:segm-robustness}
\end{table}

% \putri{While equivariant and invariant models are theoretically appealing for their robustness to geometric transformations, they can be challenging to scale \citep{mondal2023equivariant, brehmer2024does}. Foundation models, often trained on vast and varied datasets, may lack built-in equivariance, leading to performance degradation when faced with unseen geometric variations during deployment. This is examplified in \autoref{tab:segm-robustness} and \citep{mondal2023equivariant}, which demonstrate that pre-trained segmentation models may not be inherently robust to such geometric shifts. Other practical examples include the failure of 3D object recognition models struggling with non-axis-aligned point clouds \citep{vadgama2025utilityequivariancesymmetrybreaking}, or medical imaging datasets, which may be aggregated from multiple sources with differing equipment and protocols [refs]. Additionally, some works have studied the nature of learned invariance and inherently invariant architectures, and demonstrated that models trained with data augmentation, though somewhat robust, may fail under distribution shifts \citep{vadgama2025utilityequivariancesymmetrybreaking, moskalev2023genuine}.}

% \item \putri{realizing such a procedure via the \textit{canonicalization prior} \citep{mondal2023equivariant}, which is a suitable candidate due to its light-weight and post-hoc nature, fully in line with both practical and theoretical principles underlying conformal prediction.}

%% file: text/background.tex
We next provide some background on conformal prediction (\autoref{subsec:background-cp}), group equivariance and invariance properties (\autoref{subsec:background-groups}), and the canonicalization framework (\autoref{subsec:background-canon}). Regarding notation, let $\gX \times \gY$ mark the sample space with some data-generating distribution $P$ over it, and $\rvx, \rvy$ random variables with realizations $\vx, \vy$. We denote any learnable functions, such as a prediction model $f_{\vtheta}: \gX \rightarrow \gY$, as mappings with learnable parameters $\vtheta \in \Theta$.

\subsection{Conformal Prediction}
\label{subsec:background-cp}

We consider the usual setting of \emph{split conformal prediction}\footnote{As opposed to full or cross-validation conformal schemes.}, wherein a hold-out calibration set ${\gD_{cal} = \{(\vx_i, \vy_i)\}_{i=1}^{n}}$ and test set ${\gD_{test} = \{(\vx_j, \vy_j)\}_{j=n+1}^{n+m}}$ are both sampled exchangeably (\ie~permutation invariantly, see \hyperref[def:exch]{Def.~\ref{def:exch}}) from some fixed distribution $P_0$ \citep{papadopoulos2007conformal}. Using a pre-specified scoring function ${s: \gX \times \gY \rightarrow \mathbb{R}}$ and pretrained predictor $f_{\vtheta}$, we compute a set of nonconformity scores $S = \{s_i\}_{i=1}^{n}$ on $\gD_{cal}$, where ${s_i = s(f_{\vtheta}(\vx_i), \vy_i)}$. These scores encode a desired notion of disagreement between predictions and responses, such as a simple residual score $s_i = |f_{\vtheta}(\vx_i) - \vy_i|$ for regression or predicted probability $s_i = 1 - p(\rvy_i = \vy_i | \vx_i)$ for classification. Next, a sample-corrected conformal quantile $Q_{1-\alpha}(F_S)$ is computed, where $F_S$ denotes the empirical distribution over the calibration scores\footnote{Extended with $\{ +\infty \}$ to ensure proper coverage adjustments.}, and $\alpha \in (0,1)$ a tolerated miscoverage rate. Given a new test sample $(\vx_{n+1}, \vy_{n+1})$, a prediction set is then constructed as ${C(\vx_{n+1}) = \{ \vy \in \gY: s(f_{\vtheta}(\vx_{n+1}), \vy) \leq Q_{1-\alpha}(F_S) \}}$, \ie, we include candidate responses whose score does not exceed the quantile. Exploiting the data's exchangeability under $P_0$, a formal coverage guarantee on inclusion of the true response $\vy_{n+1}$ can then be given with high probability as
\begin{equation}
\label{eq:cp-guarantee}
    \mathbb{P}(\vy_{n+1} \in C(\vx_{n+1})) \geq 1 - \alpha.
\end{equation}
We refer to \cite{shafer2008tutorial, angelopoulos2024theoretical} for details on the intuition and technical proofs.

\paragraph{Mondrian conformal prediction.} The coverage guarantee in \autoref{eq:cp-guarantee} only holds \emph{marginally} over $\gD_{cal} \cup \gD_{test}$, thus ensuring coverage in a broad sense. Stronger and more refined guarantees can be obtained by simply partitioning the data into sub-populations of interest, and running the conformal procedure per partition. We refer to this as \emph{partition-conditional} or \emph{mondrian} conformal prediction \citep{toccaceli2019combination}. If we consider a mapping $\phi: \gX \times \gY \rightarrow \{1, \dots, K\}$ assigning each sample to a data partition, the coverage guarantees hold per partition as 
\begin{equation}
\label{eq:cp-guarantee-mond}
    \mathbb{P}(\vy_{n+1} \in C(\vx_{n+1}) \,|\, \phi(\vx_{n+1}, \vy_{n+1}) = k) \geq 1 - \alpha
\end{equation}
for all $k \in \{1,\dots,K\}$. Data partitions of interest can include distinction by class label \citep{cauchois2021knowing, timans2024conformalod}, feature properties \citep{m.sesia2021, c.jung2022}, or a balancing criterion like fairness \citep{y.romano2020a}.

\paragraph{Weighted conformal prediction.} To enhance data adaptivity and address settings of reduced exchangeability, a weighted formulation of CP is given by replacing the conformal quantile with $Q_{1-\alpha}(\tilde{F}_S)$, where $\tilde{F}_S$ now denotes the empirical distribution over a \emph{weighted} score set as 
\begin{equation}
\label{eq:weight-cp}
    \tilde{F}_S = \sum_{i=1}^{n} \tilde{w}_i \cdot \delta({s_i}) + \tilde{w}_{n+1} \cdot \delta({+\infty}),
\end{equation}
with $\delta({s_i})$ denoting the dirac delta centered at score $s_i$, and $\tilde{w}_i$ its associated normalized weight such that $\sum_{i=1}^{n} \tilde{w}_i = 1$. For example, \cite{barber2023conformal} suggest fixed weighting schemes such as upweighting more recent samples in a data stream setting, while \cite{guan2023localized} propose data-dependent (unnormalized) weights guided by feature distances such as the kernel distance $w_i = \exp\{-h\,|\vx_i - \vx_{n+1}|\}$.

\subsection{Group Equivariance and Invariance}
\label{subsec:background-groups}

Formally, we denote a symmetry group $G$ as a set of elements with a binary operator $\,\boldsymbol{\cdot}\,$ satisfying closure and associativity, and for which an identity element $e$ and inverses $g^{-1}$ exist such that $e \cdot g = g$ and $g^{-1} \cdot g = e$ respectively \citep{cohen2016group}. In our context, $G$ can be described as a structured space of possible symmetry transformations on the data. That is, a sample $\rvx \in \gX$ is transformed by a \emph{group action} as $\rho(g) \cdot \rvx$, where $g \in G$ denotes a group element and $\rho: G \rightarrow T$ a group representation mapping $g$ to a concrete transformation\footnote{$T \subset GL(V)$ denotes a subset of the total set of linear invertible transformations on some vector space $V$.}. For instance, if we define $G = SO(2)$ as the group of planar rotations, then $g$ might represent a particular rotation angle, and $\rho(g)$ the rotation of $\rvx$ by that angle via matrix multiplication. Given such geometric data transformations, desirable properties for some predictor $f_{\vtheta}$ can include \emph{(i)} preserving the symmetry structure of $G$ by commuting with group actions, \ie~being \emph{equivariant}, or \emph{(ii)} ensuring robustness to group actions by remaining \emph{invariant} to them. Specifically, $f_{\vtheta}$ is deemed group equivariant if for all $g \in G$ we have that
\begin{equation} 
    f_{\vtheta}(\rho(g) \cdot \rvx) = \rho'(g) \cdot f_{\vtheta}(\rvx),
\label{eq:equivariance}
\end{equation}
where $\rho(g)$ and $\rho'(g)$ act on the data input space $\gX$ and output space $\gY$, respectively. Thus the model's output commutes predictably with the applied transformation, a property frequently employed, for instance, in translation-equivariant convolutional models for image processing. In contrast, if $\rho'(g) = \mathbb{I}$ equates the identity transformation for any group element $g$, then $f_{\vtheta}$ is group-invariant to $G$. This property is desirable if input samples $\rvx$ are subject to geometric data transformations or shifts, but we desire $f_{\vtheta}$ to provide consistent prediction outputs regardless. In neural network models, both properties are typically achieved by employing architectures that inherently incorporate \autoref{eq:equivariance} as a constraint, or through explicit or implicit learning of symmetries, \eg~via data augmentation (see \autoref{sec:related-work}).

\subsection{Equivariance via Canonicalization}
\label{subsec:background-canon}

Instead of designing a model and its layers to be equivariant, one may also obtain equivariance through \emph{canonicalization} \citep{mondal2023equivariant, kaba23equivariance}. At its core, canonicalization aims to learn a mapping from potentially transformed data to its standardized or canonical orientation before processing by the predictor. The approach separates the tasks of correcting and predicting for transformed data, greatly increasing flexibility by allowing the use of \emph{non}-equivariant pretrained predictors within an equivariant framework. More formally, given a predictor $f_{\vtheta}$ we additionally consider a learnable \emph{canonicalization network} (CN) as $c_{\vtheta}: \gX \rightarrow G$, and denote the canonicalization process as 
\begin{equation}
f_{\vtheta}(\rvx) = \rho'(c_{\vtheta}(\rvx)) \cdot f_{\vtheta}\left(\rho(c_{\vtheta}(\rvx)^{-1}) \cdot \rvx \right).
\label{eq:canon-equiv}
\end{equation}
The CN $c_{\vtheta}$ aims to predict the (inverse) group element to map $\rvx$ back to its canonical form, and \autoref{eq:canon-equiv} ensures $f_{\vtheta}$ is $G-$equivariant if $c_{\vtheta}$ itself is $G-$equivariant \citep{kaba23equivariance}. Similarly, for invariance we have $\rho'(g) = \mathbb{I}$ and \autoref{eq:canon-equiv} simplifies to 
\begin{equation}
f_{\vtheta}(\rvx) = f_{\vtheta}\left(\rho(c_{\vtheta}(\rvx)^{-1}) \cdot \rvx \right) = f_{\vtheta}\left(\hat{g}^{-1} \cdot \rvx \right),
\label{eq:canon-inv}
\end{equation}
where we've omitted $\rho$ since there is no ambiguity on the group action space\footnote{And subsequently abuse notation for simplicity and use $g \cdot \rvx$ as the application of $g$ on the domain directly.}, and $\hat{g}$ denotes the predicted group element using $c_{\vtheta}$. Whereas the original formulation by \cite{kaba23equivariance} directly predicts a single group element ${\hat{g} = c_{\vtheta}(\rvx)}$, \cite{mondal2023equivariant} extend the approach to predict a group distribution $\hat{P}_{G\mid\rvx}$ over transformations, in which case ${\hat{g} \sim \hat{P}_{G|\rvx}}$ can be sampled.

\paragraph{Regularization using the canonicalization prior.} There are practical challenges in ensuring that the learning process of the CN is both coupled to the employed predictor and the correct poses in the data. Thus, \cite{mondal2023equivariant} propose training the CN with a double objective of the form ${\mathcal{L}_{\text{total}} = \mathcal{L}_{\text{task}} + \beta \cdot \mathcal{L}_{\text{prior}}}$, where $\mathcal{L}_{\text{task}}$ is a cross-entropy loss term and $\mathcal{L}_{\text{prior}}$ a regularization term. In particular, if $f_{\vtheta}$ is pretrained and fully frozen during training the task loss is zero, and an additional learning signal becomes necessary. Thus, the \emph{canonicalization prior} (CP) term $\mathcal{L}_{\text{prior}}$ is introduced to align the CN's learned poses with the canonical pose prevalent in the data $\gD_{can} \sim P_{can}$ used to learn the CN (\eg~a hold-out data split). The loss is then given by 
\begin{equation}
    \mathcal{L}_{\text{prior}} = \mathbb{E}_{P_{can}}[D_{KL}(P_{G \mid \rvx} \;||\; \hat{P}_{G \mid \rvx})],
\label{eq:canon-prior}
\end{equation}
where $P_{G \mid \rvx}$ is a prior distribution for the group elements acting on samples in $\gD_{can}$, and $D_{KL}$ the Kullback-Leibler divergence. In practice the prior is usually set to $P_{G \mid x} = \delta(e)$, \ie, full probability mass on the identity element, thus assuming the `correct' data is subject to no transformations. This additionally simplifies computation of \autoref{eq:canon-prior} for particular groups, \eg~for discrete rotations we obtain $\mathcal{L}_{\text{prior}} = - \mathbb{E}_{P_{can}} \log \hat{P}_{G \mid \rvx}(e)$, the negative log probability of the identity element \citep{mondal2023equivariant}. Note that $G$ still needs to be defined beforehand, \ie~the CN learns a distribution \emph{over} group elements, rather than a set of valid group elements themselves (from a possibly infinite space). However, we find that results are not overly impacted when the correct group is a subgroup $G' \subset G$ of the model-specified group (\eg, $C4$ rotations rather than $C8$ rotations), providing some leeway to misspecification (see \autoref{tab:cifar100-robust}).

%% file: text/method.tex
We next motivate why canonicalization suits itself naturally for joint use with conformal prediction, including a perspective on data exchangeability. In \autoref{subsec:method-usecases} we then outline three ways to leverage obtained geometric information for conformal procedures under differing shift scenarios.

\paragraph{Practical motivation: flexible and efficient.} Equivariance modelling usually requires custom prediction models which embed the necessary geometric constraints deep within their architecture, such as via group convolutions with regular \citep{cohen2016group,Bekkers2020B-Spline} or steerable filters \citep{weiler2019general}. This introduces additional complexity into the model, complicates training, and can hamper the transferability of a solution across datasets or tasks. In contrast, canonicalization effectively decouples the prediction and equivariance components, permitting the use of a broader variety of non-equivariant, pretrained models for prediction, and ensuring equivariance in a \emph{post-hoc} step. This outsourcing permits the use of more efficient, light-weight equivariant models to learn the canonical mapping in an unsupervised way, while the complex prediction task is handled by a separate, usually substantially larger model (magnitudes larger, see \autoref{subsec:exp-robust}). This can also provide benefits over data augmentation, since only a single forward pass through the predictor is necessary. Most crucially, the obtained flexibility meshes particularly well with the conformal prediction framework, as CP's key advantage of \emph{post-hoc} compatibility with arbitrary `black-box' predictors is preserved. In that sense, we may think of canonicalization as a second `bolt-on' module, situated inbetween the predictor and uncertainty estimation via CP. Naturally, canonicalization has little to no effect on models that are \emph{already} symmetry-aware, as the additional module then becomes redundant.

\paragraph{Theoretical motivation: canonical mapping as data exchangeability.} We may also motivate canonicalization for CP from a more fundamental data perspective. Intuitively, the canonicalization network $c_{\vtheta}$ aids mitigate the predictor's performance loss due to encountered geometric shifts by enforcing data exchangeability with the training set, in turn benefitting uncertainty estimation. More formally, let us first define \emph{data exchangeability} following the CP framework:
\begin{definition}[Exchangeability \citep{shafer2008tutorial}]
    A sequence of random variables $\rvx_1, \dots, \rvx_n$ is exchangeable if for any permutation $\pi: \{1, \dots, n\} \rightarrow \{1, \dots, n\}$ with $n \geq 1$ we have that $P(\rvx_1, \dots, \rvx_n) = P(\rvx_{\pi(1)}, \dots, \rvx_{\pi(n)})$.
\vspace{-5mm}
\label{def:exch}
\end{definition}
That is, the joint data probability is invariant to sample ordering. In particular, observe how the \emph{i.i.d} setting is a special case where $P(\rvx_1, \dots, \rvx_n)$ factorizes. For conformal coverage guarantees along \autoref{eq:cp-guarantee} to nominally hold, \hyperref[def:exch]{Def.~\ref{def:exch}} only needs to be satisfied across calibration ($\gD_{cal}$) and test data ($\gD_{test}$), but \emph{not} necessarily for the predictor's training set ($\gD_{train}$). However, learned data properties that poorly translate to new (shifted) samples will result in a low-quality set of computed nonconformity scores, starkly inflating prediction set sizes and rendering obtained sets uninformative. From a data perspective, this issue can be alleviated if $\gD_{train}$ also approximately satisfies \hyperref[def:exch]{Def.~\ref{def:exch}}, and $f_{\vtheta}$ thus guarantees informative scoring. 

This is precisely what the CN attempts to ensure via its canonical mapping. Classical exchangeability imposes data invariance under permutations $\pi \in \mathbb{S}_n$, where $\mathbb{S}_n$ represents the set of all permutations in $\{1,\dots,n\}$. Assuming a shift by the group $G$ affecting $\gD_{cal}$, each calibration sample $\rvx_i$ is now also susceptible to an independent transformation ${g_i \in G}$. That is, on a dataset level we now aim for exchangability (\ie~group invariance) to extend to the group $G^n = G \times G \times \dots \times G$, in which each sample experiences a potentially different transformation of $G$. For every affected sample $g_i \cdot \rvx_i$, the CN ensures the existence of an inverse transform $c_{\vtheta}(g_i \cdot \rvx_i)^{-1}$ which neutralizes $g_i$. That is, proper canonicalization maintains the relationship ${c_{\vtheta}(g \cdot \rvx)^{-1} = c_{\vtheta}(\rvx)^{-1} \cdot g^{-1}}$ for all $g \in G$ and inputs $\rvx$ \citep{kaba23equivariance}. Under the action of $G^n$, we then observe for the joint distribution that
\begin{multline*}
    P(c_{\vtheta}(g_1 \cdot \rvx_1)^{-1} \cdot g_1 \cdot \rvx_1, \,\dots\,, c_{\vtheta}(g_n \cdot \rvx_n)^{-1} \cdot g_n \cdot \rvx_n) \\
    = P(c_{\vtheta}(\rvx_1)^{-1} \cdot \rvx_1, \,\dots\,, c_{\vtheta}(\rvx_n)^{-1} \cdot \rvx_n) \, ,
\end{multline*}
ensuring that the distribution over canonicalized samples remains invariant under $G^n$. This generalizes the classical exchangeability definition of \hyperref[def:exch]{Def.~\ref{def:exch}} to include both dataset permutations and sample-wise transformations, enlarging the symmetry group from $\mathbb{S}_n$ to $\mathbb{S}_n \times G^n$. Since distributional invariance to $G$ implies that transformations from $G$ do not alter the joint distribution, the CN effectively enforces \textit{probabilistic symmetry} (\cite{bloem2020probabilistic}, Prop. 1). Thus, it guarantees well-calibrated nonconformity scores practically useful for CP even under geometric shifts.

\subsection{Use Cases for Conformal Prediction}
\label{subsec:method-usecases}

Following our motivation, we now illustrate three interesting ways how obtained group information can be leveraged to benefit different conformal prediction procedures and tasks.

\input{fig/fig_cond_group_dist}

\paragraph{For general robustness to geometric data shifts.} We first directly demonstrate the obtained robustness to a geometric data shift at calibration and test time. To that end, we can simply combine the CN $c_{\vtheta}$ with a non-equivariant, pretrained predictor and apply standard split conformal prediction (SCP). Since the CN ensures the necessary exchangeable mapping to align the predictor's outputs with the conformal procedure, we expect a substantial improvement in prediction set sizes over directly using $f_{\vtheta}$ and SCP without canonicalization. 

\paragraph{As a diagnostics tool and proxy for conditional coverage.} Unlike inherently equivariant models or models trained with data augmentation, canonicalization provides us with explicit access to \emph{sample-wise} geometric information or pose via the group distributions $\hat{P}_{G \mid \rvx}$. These can be exploited to construct empirical group distributions pertaining to any separable data partition of interest, \eg~by class labels or feature properties. Such empirical group distributions can provide insights into the geometric poses under which a certain property or partition naturally occurs in the data (see \autoref{app:details-mcp} for further intution). If a partition's `group map'---visualized for some examined partitions in \autoref{fig:class_conditional_mondrian}---reveals informative geometric patterns, the group assignments can be subsequently leveraged to provide stronger partition-conditional or mondrian conformal prediction (MCP) guarantees (\autoref{eq:cp-guarantee-mond}). In that sense, the group information can be leveraged as a data diagnostics tool to uncover even \emph{a priori} unknown but geometrically informative partitions, or to suggest data exchangeability for a partition when no meaningful group pattern emerges. In principle, such group maps could be extended as far as incorporating multiple datasets to potentially uncover geometric shifts across new data sources.

Formally, given some data partition $k \in \{1, \dots, K\}$ of $\gD_{cal}$ into $K$ parts, an empirical group distribution for the $k$-th partition can be constructed as $\hat{P}_{G \mid k} = \{ \hat{P}_{g \mid k} \mid g \in G\}$, where $\hat{P}_{g \mid k}$ denotes the $g$-th element's estimated frequency computed as
\begin{align}
    \hat{P}_{g \mid k} &= \frac{\sum_{i=1}^{n}\mathbbm{1}(\hat{g}_i = g \,\wedge\, \phi(\vx_i, \vy_i) = k) }{\sum_{i=1}^{n}\mathbbm{1}(\phi(\vx_i, \vy_i) = k) }.
    \label{eq:conditional_distribution}
\end{align}
The indicator function is given by $\mathbbm{1}[\cdot]$, whereas $\hat{g}_i \sim \hat{P}_{G | \vx_i}$ is the sampled group element obtained for $(\vx_i, \vy_i)$. 

\paragraph{As a weighting scheme for double shift settings.} Consider a more complex \emph{double shift} setting, wherein the first shift between $\gD_{train}$ and $\gD_{cal}$ is addressed by the CN, but an additional second shift between $\gD_{cal}$ and $\gD_{test}$ occurs. For example, the CN trained on $\gD_{cal}$ learns to address a shift caused by the $C8$ rotation group, but test samples are susceptible to continuous rotations on $SO(2)$. In this case even the use of canonicalization with standard SCP can be insufficient to ensure conformal guarantees, since the CN can underperform when faced with new, unknown group elements (\ie~rotation angles in $SO(2)$ but not $C8$). However, the obtained group information can still be leveraged to inform \emph{geometric weights} for a weighted conformal prediction strategy (WCP). We posit that the CN assigns higher probability to group elements that are `closer' aligned with the test sample's unknown transformation, and as such provides information to upweigh more geometrically relevant calibration samples; we elaborate on this intuition in \autoref{app:details-wcp}. In conjunction with WCP, this may offer improved robustness against shifts with \emph{unknown} group elements. Different double-shift settings and approaches to establish robustness are outlined in \autoref{tab:wcp-settings}\footnote{We empirically examine the first (\autoref{subsec:exp-robust}) and last rows (\autoref{subsec:exp-weightcp}).}.

\input{tab/double_shift_explainer}

\input{tab/robust_shift_cifar100_aps}

Formally, given a test instance $\vx_{n+1}$, the $i$-th calibration sample's geometric relevance with respect to $\vx_{n+1}$ can be measured by ${D \bigl( \hat{P}_{G \mid \vx_{n+1}}, \hat{P}_{G \mid \vx_i} \bigl)}$, with $D$ being any distributional distance metric. Since we desire a small geometric distance between two samples to produce a large importance weight, the unnormalized weight $w_i$ can be defined by an inverse relation of the form $w_i(\vx_{n+1}) = 1/(1 + D^p)$, where $p$ denotes an additional parameter modulating the slope or skewdness of the weighting distribution. A final weight $\tilde{w}_i$ is then acquired by subsequent normalization.

% Note that we desire the geometric distance between two samples to be \emph{small} for a large weight, so we might want to take the inverse $1/d$ or a related idea. Finally, we need weights to be normalized for their use with WCP, \eg~by using sum normalization $\tilde{w}_i(x_{n+1}) = \frac{w_i(x_{n+1})}{\sum_{j=1}^n w_j(x_{n+1})}$ or softmax normalization.

% , or with $\hat{g} \sim \hat{P}_{G\mid x}$ as $d\bigl(\hat{g}_{n+1}, \hat{P}_{G\mid x_i} \bigl)$ or even $d\bigl(\hat{g}_{n+1}, \hat{g}_{i} \bigl)$ more coarsely

%% file: fig/fig_cond_group_dist.tex
\begin{figure*}[th]
    \centering
    % First row of figures
    \begin{subfigure}{0.2\textwidth}
        \centering
        \includegraphics[width=\linewidth]{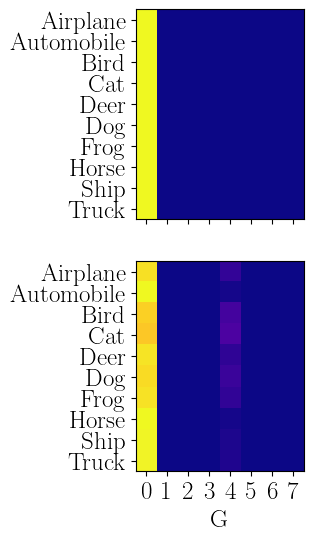}
        % \caption{Subcaption for image 1}
    \end{subfigure}
    % \hfill
    \begin{subfigure}{0.2\textwidth}
        \centering
        \includegraphics[width=\linewidth]{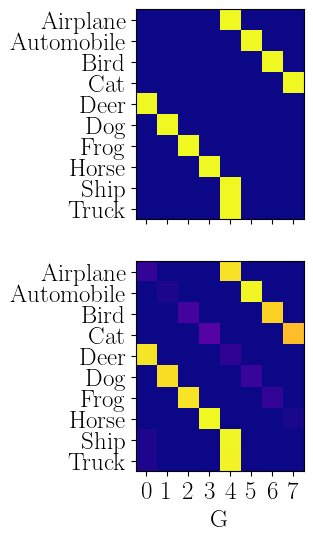}
        % \caption{Subcaption for image 2}
    \end{subfigure}
        % \hfill
    \begin{subfigure}{0.251\textwidth}
        \centering
        \includegraphics[width=\linewidth]{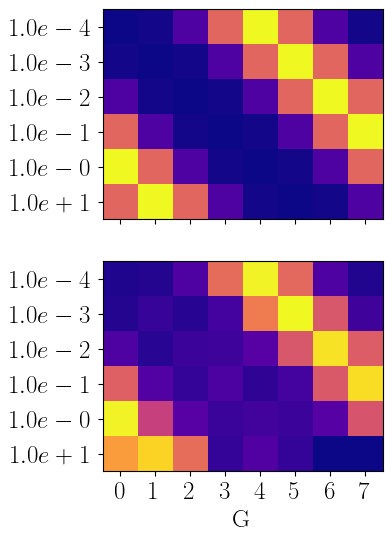}
        % \caption{Subcaption for image 2}
    \end{subfigure}
    % \hfill
    \begin{subfigure}{0.205\textwidth}
        \centering
        \includegraphics[width=\linewidth]{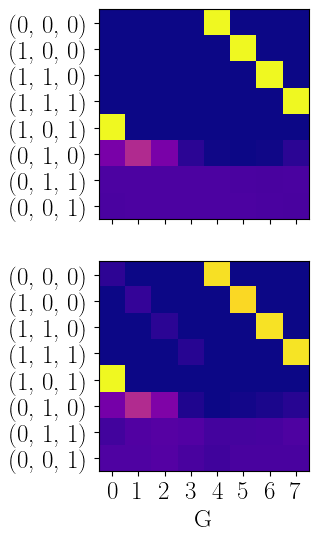}
        % \caption{Subcaption for image 2}
    \end{subfigure}

    \vspace{-3mm}
    % \vspace{1em} % Add vertical space between rows

    \caption{True (\emph{top}) and recovered (\emph{bottom}) partition-conditional group distributions based on different data partitions (class, entropy, color) and shifts (outlined in \autoref{app:details-mcp}). \emph{From left to right:} (a) Class partition, no shift; (b) Class partition, \texttt{dirac} shift; (c) Entropy partition, \texttt{normal} shift; (d) Color partition, \texttt{var-gauss} shift. For (c) samples are partitioned by predictive entropy into exp. scaled bins, for (d) samples are assigned based on RGB values and average image color. Both are used as a measure of sample complexity. We consider $G$ as the $C8$ rotation group. Using the canonicalization network's group distributions and \autoref{eq:conditional_distribution}, the built group maps (\emph{bottom}) accurately uncover existing partition-conditional geometric patterns.
    }
    \label{fig:class_conditional_mondrian}
\end{figure*}

%% file: tab/double_shift_explainer.tex
\begin{table}[!t]
    \caption{Different possible geometric shift settings for calibration and test data. Note that the CN is trained on $\gD_{cal}$ and thus learns mappings to $G$. The density ratio $w(\rvx)$ denotes reweighting on the same group support, while $G_{new}$ denotes a group with newly encountered group elements, \ie~$G \subset G_{new}$. The train data for the predictor $f_{\vtheta}$ in all cases is unaffected by geometric shift, \ie~$G_{train} = \{e\}$.}
    \centering
    \begin{tabular}{cccl}
        \hline
        \textbf{Train} & \textbf{Calibration} & \textbf{Test} & \textbf{Robustness}\\
        \hline
        $\delta(e)$ & $P_{G \mid \rvx}$ & $P_{G \mid \rvx}$ & CN + SCP \\
        $\delta(e)$ & $P_{G \mid \rvx}$ & $w(\rvx) \cdot P_{G \mid \rvx}$ & CN + SCP \\
        $\delta(e)$ & $P_{G \mid \rvx}$ & $P_{G_{new} \mid \rvx}$ & CN + WCP \\
        \hline
    \end{tabular}
    \label{tab:wcp-settings}
\end{table}

%% file: tab/robust_shift_cifar100_aps.tex
\begin{table*}[!t]
  \caption{
    Results with APS on CIFAR-100 for target coverage $(1-\alpha) = 95\%$ across different rotation shifts. Target coverage is efficiently maintained when no shift occurs, but grows excessively large and uninformative (\colorbox{gray!25}{\textcolor{gray!25}{oo}}) when the underlying predictor $\hat{f}_{\vtheta}$ is not equivariant, or if the wrong group is specified (learning $C4$ but exposed to $C8$). Results are reported across $T=10$ random calibration/test splits. \# Param. indicates the required number of training parameters. CP$^2$ employs the same (pretrained and frozen) predictor $\hat{f}_{\theta}$, and only requires training a substantially smaller canonicalization network.
  }
  \centerline{
  \setlength{\tabcolsep}{7pt}
  \resizebox{\linewidth}{!}{
      \begin{tabular}{ll|ccc|ccc|ccc}
        \toprule
         & & \multicolumn{3}{c}{\textbf{No Shift}} & \multicolumn{3}{c}{\textbf{$C4$ Rotation Shift}}  & \multicolumn{3}{c}{\textbf{$C8$ Rotation Shift}} \\
         \textbf{Model} & \# Param. & Acc & Coverage & Set Size & Acc & Coverage & Set Size & Acc & Coverage & Set Size \\
         \midrule
         $\hat{f}_{\theta}$ & 23.7 M
         & 71.66 & $ 95.09 \pm .003 $ & $ 6.212 \pm .106 $ 
         & 40.17 & $ 96.49 \pm .002 $ & \cellcolor{gray!25}$ 55.149 \pm .388 $ 
         & 33.68 & $ 95.63 \pm .001 $ & \cellcolor{gray!25}$ 61.402 \pm .524 $ 
         \\
         
         $\hat{f}_{\theta}$ + $SO(2)$ Aug. & 23.7 M
         & 60.13 & $ 95.02 \pm .005 $ & $ 11.362 \pm .393 $ 
         & 59.72 & $ 95.20 \pm .003 $ & $ 11.634 \pm .302 $ 
         & 58.27 & $ 95.00 \pm .004 $ & $ 11.213 \pm .320 $ \\
         
         $\hat{f}_{\theta}$ + $C_4$ Aug. & 23.7 M
         & 63.03 & $ 95.05 \pm .006 $ & $ 10.047 \pm .475 $ 
         & 62.72 & $ 95.11 \pm .005 $ & $ 10.175 \pm .452 $ 
         & 49.72 & $ 98.10 \pm .001 $ & \cellcolor{gray!25}$ 88.892 \pm .249 $ \\
         $\hat{f}_{\theta}$ + $C_8$ Aug. & 23.7 M
         & 62.53 & $ 95.38 \pm .005 $ & $ 10.758 \pm .361 $ 
         & 62.37 & $ 95.33 \pm .002 $ & $ 10.396 \pm .183 $ 
         & 60.82 & $ 95.26 \pm .002 $ & $ 10.970 \pm .156 $ \\
         \midrule
         CP$^2$ ($G$ = 4)   & 1.0 M
         & 65.37 & $ 95.05 \pm .004 $ & $ 10.583 \pm .431 $ 
         & 65.37 & $ 95.02 \pm .004 $ & $ 10.557 \pm .443 $
         & 48.19 & $ 95.02 \pm .005 $ & \cellcolor{gray!25}$ 32.605 \pm 2.382 $ \\
         
         CP$^2$ ($G$ = 8)  & 2.0 M
         & 65.46 & $ 95.34 \pm .004 $ & $ 11.198 \pm .398 $ 
         & 65.46 & $ 94.95 \pm .004 $ & $ 10.820 \pm .406 $ 
         & 63.94 & $ 94.81 \pm .003 $ & $ 11.296 \pm .396 $ \\
        \bottomrule
      \end{tabular}
      }
  }
  \label{tab:cifar100-robust}
\end{table*}

%% file: text/exp.tex
We next empirically validate our three different approaches to integrating geometric information with conformal procedures. We briefly outline our experiment design, with further details and results in the Appendix. Our code is publicly available at \url{https://github.com/computri/geometric_cp}.

\paragraph{Experimental design.} As outlined in \autoref{subsec:background-canon}, the canonicalizer is usually trained using a joint task and prior regularization loss ${\mathcal{L}_{\text{total}} = \mathcal{L}_{\text{task}} + \beta \cdot \mathcal{L}_{\text{prior}}}$. In accordance with conformal prediction we desire a fully \emph{post-hoc} approach amenable to pretrained predictors, and as such leverage canonicalizers trained exclusively with the canonicalization prior via \autoref{eq:canon-prior}. Consequently, the predictor in our experiments is \emph{pretrained and frozen} when used in conjunction with the CN, whereas relevant data augmentation and equivariant baselines require $\hat{f}_\theta$ to be trained from scratch. We emphasize that our approach leverages the exact same prediction model $\hat{f}_\theta$, pretrained without augmentations. 

Given various classification tasks, we employ the popular \textit{Adaptive Predictive Sets} (APS) \citep{romano2020classificationvalidadaptivecoverage} as our default nonconformity scoring approach for any conformal procedure, and report results for an alternative scoring method by \cite{sadinle2019least} in \autoref{app:exp}. Following standard practice we report \emph{empirical coverage} and \emph{mean set size} as our metrics to assess the quality of uncertainty estimates \citep{shafer2008tutorial, angelopoulos2024theoretical}. Empirical coverage determines the \emph{validity} of our guarantees by comparing to the target coverage level $(1-\alpha)$, whereas prediction set sizes assess the \emph{efficiency} of the method, and lower set sizes are more informative. We dub the proposed approach CP$^2$, for the combined use of the canonicalization prior (CP) with conformal prediction.

\subsection{Robustness to Geometric Data Shifts}
\label{subsec:exp-robust}

We assess the method's robustness to three geometric shifts caused by $C4$, $C8$, and $SO(3)$ rotation groups, and across three datasets (CIFAR-10, CIFAR-100, and ModelNet-40) and two data modalities (images and point clouds).

\input{tab/robust_shift_pointcloud_aps}

\paragraph{Image classification.} We evaluate two ResNet-50 predictors on CIFAR-10 and CIFAR-100 samples subjected to $C4$ and $C8$ rotation shifts. These groups form discretized subgroups of $SO(2)$ with four and eight equidistant elements, respectively. Three model training configurations are considered: \emph{(i)} the prediction models trained in a default, non-augmented manner; \emph{(ii)} the same predictors trained with relevant data augmentations to obtain approximate invariance; and \emph{(iii)} pretrained and frozen predictors with `bolt-on' canonicalization models trained for $G=4$ and $G=8$ group elements. Each configuration is subsequently combined with standard SCP to provide prediction sets with a target coverage rate of $(1-\alpha)=95\%$.

Classification accuracy and conformal results for CIFAR-100 are given in \autoref{tab:cifar100-robust} (see \autoref{tab:cifar10-robust} for CIFAR-10). For non-shifted data, performance remains comparable. While the base predictor exhibits highest accuracy in that setting, it lacks generalizability under geometric shift, reflected by its poor performance and uninformative set sizes. In contrast, both data-augmented and canonicalization approaches ensure robustness to the shift, while achieving similar accuracy as in the non-shifted setting unless the learned group is misspecified (\ie, trained for $C4$ but exposed to $C8$). We observe that in the inverse case robustness continues to hold, thus suggesting to favour a broader group definition when faced with the risk of unknown group elements. That is, ideally the learned group is chosen to be maximal within constraints on computational resources and accuracy requirements, since a coarser discretization will induce more discretization artifacts. Overall, our results highlight canonicalization as a light-weight alternative to ensure robustness without necessitating retraining. 

\paragraph{Point cloud classification.} Unlike 2D images, point clouds exist within a continuous 3D space where rotational shifts are more intrinsic. We evaluate the performance of popular point cloud classifiers PointNet \citep{qi2016pointnet} and DGCNN \citep{wang2018dgcnn} with and without canonicalization, along with Rapidash \citep{vadgama2025utilityequivariancesymmetrybreaking}, a recent proposal which permits adjustable levels of equivariance---from non-equivariant to fully equivariant. Our results in \autoref{tab:pointcloud-robust} echo those from the image domain, revealing that unadjusted base models fail to maintain robustness against orientation shifts in point clouds, resulting in inflated conformal metrics. Conversely, models equipped with data augmentation or equivariance properties demonstrate better resilience to these geometric shifts. In particular this includes canonicalization, which in this particular instance trains a network by \emph{multiple magnitudes} smaller than other approaches (see \autoref{app:details-robust} for architecture details). In addition, data augmentations become substantially more expensive to incorporate due to the high degrees of freedom offered by 3D spatial rotations.

\input{fig/fig_mcp_cov_one}

\subsection{Diagnostics for Conditional Coverage}
\label{subsec:exp-condcover}

Next, we leverage the geometric information obtained from the canonicalization network's sample-wise group distributions to construct partition-conditional group distributions $\hat{P}_{G|k}$ following \autoref{eq:conditional_distribution}, and visualize the obtained `group maps' for CIFAR-10 in \autoref{fig:class_conditional_mondrian}. In each column, we display the true group distribution $P_{G|k}$---tractable by manually inducing different partition-conditional shifts---and the recovered distribution $\hat{P}_{G|k}$ using the CN. Indeed, we find that the model can effectively uncover meaningful geometric patterns when particular shifts are imbued on the data. We also visualize the class-conditional group map on the data \emph{without} any geometric shift (\autoref{fig:class_conditional_mondrian}, first column) and observe how samples across all classes are predominantly mapped into the identity element, \ie~upright. We can interpret the approach as a visual test for exchangeability, assessing whether all bucketed samples across the partition adhere to the same geometric properties (as is the case here).

\input{fig/fig_wcp}

Additionally, we may determine that particular partitions correlate with particular group elements, and in such cases leverage sample assignments to each entry $\hat{P}_{g \mid k}$ as an unsupervised proxy for mondrian conformal prediction. While we conformalize directly for the group partition (see \autoref{fig:app-partition_coverage}), the captured geometric relationship will \emph{by proxy} lead to more balanced coverage for the associated data partition. We demonstrate this for the class-conditional case in \autoref{fig:partition_coverage}, where MCP applied to the $C8$ group elements---which exhibit a correspondence with CIFAR-10 class labels---substantially improves per-class coverage over SCP due to a better-tailored conformal quantile estimate. Naturally, the proxy relationship is limited by the extent to which a partition-conditional group pattern intersects with multiple partitions, and at the other end no improvements are obtained when partitions exhibit identical group distributions (\autoref{fig:target_partition_coverage}, left). 

\subsection{Weighting for Double Shift Settings}
\label{subsec:exp-weightcp}

Finally, we evaluate the double shift setting described in \autoref{subsec:method-usecases} and \autoref{tab:wcp-settings} (third row). \autoref{fig:weighting} depicts the encounter of $C4$ and $C8$ discrete rotation shifts on $\gD_{cal}$, with a gradual interpolation for different continuous $SO(2)$ shifts (by adjusting data sampling probabilities) on $\gD_{test}$. Thus, the secondary test shift ranges from benign (\ie~no new group elements) to severe, where the uniform $SO(2)$ group places much probability mass on previously unknown rotations. We employ our geometric weighting scheme in conjunction with weighted conformal prediction, and compare against the standard variant. We clearly observe a coverage breakdown with growing geometric difference between data partitions, since the necessary exchangeability condition between $\gD_{cal}$ and $\gD_{test}$ is invalidated. Yet, geometric weighting can help delay the effect at the cost of enlarged set sizes, suggesting partial group knowledge can be beneficial to robustness even under \emph{unkown} group actions. However, improvements remain bottlenecked by the static training performance of the canonicalizer, and a more practical deployment should consider an updating step to incorporate new geometric information upon arrival.

%% file: tab/robust_shift_pointcloud_aps.tex
\begin{table*}[!t]
  \caption{
  Results with APS on ModelNet-40 for target coverage $(1-\alpha) = 95\%$ across $SO(3)$ rotation shift. Target coverage is efficiently maintained when no shift occurs, but grows excessively large and uninformative (\colorbox{gray!25}{\textcolor{gray!25}{oo}}) when the underlying predictor $\hat{f}_{\vtheta}$ is not equivariant. Results are reported across $T=10$ random calibration/test splits. \# Param. indicates the required number of training parameters. (*) denotes the use of a slightly different data preprocessing and training split.
  }
  \centerline{
  \setlength{\tabcolsep}{7pt}
  \resizebox{\linewidth}{!}{
      \begin{tabular}{ll|ccc|ccc}
        \toprule
         & & \multicolumn{3}{c}{\textbf{No Shift}} & \multicolumn{3}{c}{\textbf{$SO(3)$ Rotation Shift}}  \\
         \textbf{Model} & \# Param. & Acc & Coverage & Set Size & Acc & Coverage & Set Size \\
         \midrule
         PointNet & 0.7 M
         & 87.49 & $ 94.93 \pm .008 $ & $ 2.133 \pm .079 $ 
         & 8.73  & $ 98.01 \pm .003 $ & \cellcolor{gray!25} $ 38.835 \pm .062 $ \\
         DGCNN & 1.8 M
         & 91.41 & $ 94.76 \pm .007 $ & $ 1.36 \pm .055 $ 
         & 15.15 & $ 95.62 \pm .004 $ & \cellcolor{gray!25} $ 36.218 \pm .161 $ \\
         % 2
         Rapidash* & 1.7 M
         & 86.51 & $ 95.34 \pm .007 $ & $ 1.540 \pm .073 $ 
         & 12.84 & $ 99.95 \pm .000 $ & \cellcolor{gray!25} $ 39.956  \pm .018 $ \\
         \midrule
         PointNet + $SO(3)$ &  0.7 M
         & 57.46 & $ 95.3 \pm .008 $ & $ 6.532 \pm .384 $ 
         & 55.63 & $ 94.85 \pm .011 $ & $ 7.058 \pm .497 $ \\
         DGCNN + $SO(3)$ & 1.8 M
         & 86.55 & $ 95.18 \pm .004 $ & $ 1.508 \pm .040 $ 
         & 85.74 & $ 94.71 \pm .010 $ & $ 1.554 \pm .105 $ \\
         % 3
         Rapidash* + $SO(3)$ & 1.7 M
         & 76.18 & $ 94.99 \pm .006 $ & $ 3.630 \pm .181 $ 
         & 76.70 & $ 94.72 \pm .005 $ & $ 3.566 \pm .151 $ \\
         % \midrule
         % 5
         % Invariant Rapidash* & 
         % & 73.66 & $ 94.48 \pm .007 $ & $ 5.174 \pm .293 $ 
         % & 73.50 & $ 95.13 \pm .007 $ & $ 4.908 \pm .216 $ \\
         % % 1
         Invariant Rapidash* + $SO(3)$  & 1.7 M
         & 74.39 & $ 95.24 \pm .003 $ & $ 4.747 \pm .142 $ 
         & 74.02 & $ 95.26 \pm .007 $ & $ 5.255 \pm .348 $ \\
         % 6
         % Equivariant Rapidash* & 
         % & 87.56 & $ 95.06 \pm .007 $ & $ 1.466 \pm .079 $ 
         % & 22.37 & $ 96.84 \pm .026 $ & $ 37.04 \pm 2.406 $ \\
         % % 4
         Equivariant Rapidash* + $SO(3)$ & 2.0 M
         & 88.70 & $ 94.82 \pm .006 $ & $ 1.427 \pm .040 $ 
         & 87.68 & $ 94.94 \pm .007$ & $ 1.438 \pm .067 $ \\
         \midrule
         CP$^2$ (PointNet) & 1.3 K
         & 62.11 & $ 95.47 \pm .008 $ & $ 6.912 \pm .230 $ 
         & 62.07 & $ 94.94 \pm .007 $ & $ 6.604 \pm .226 $ \\
         CP$^2$ (DGCNN) & 1.3 K
         & 85.78 & $ 95.11 \pm .007 $ & $ 2.317 \pm .112 $ 
         & 85.86 & $ 94.46 \pm .008 $ & $ 2.262 \pm .105 $ \\
        \bottomrule
      \end{tabular}
      }
  }
  \label{tab:pointcloud-robust}
\end{table*}

%% file: fig/fig_mcp_cov_one.tex
\begin{figure}[!t]
    \centering
    \includegraphics[width=\linewidth]{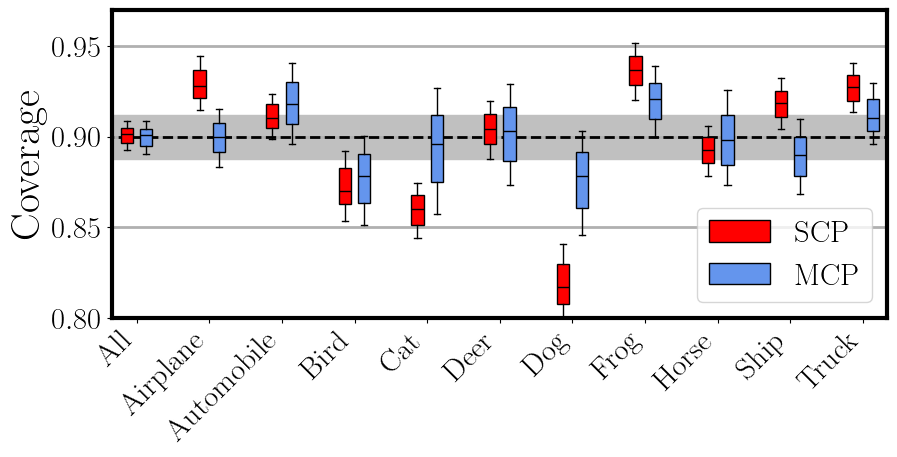}
    % \caption{Weighting mechanism 2.}
    \label{fig:subfig2}
    % \vspace{-5mm}
\caption{
    Per-class coverage results for split (SCP) and mondrian conformal prediction (MCP) on CIFAR-10, for a target coverage of $(1-\alpha)=90\%$. Leveraging MCP on the group partition, improved coverage balance is obtained \emph{by proxy} for the class partition due to an accurately captured geometric relationship between class labels and $C8$ group elements by the group maps (\autoref{fig:class_conditional_mondrian}, second column).
}
\label{fig:partition_coverage}
\end{figure}

%% file: fig/fig_wcp.tex
\begin{figure*}[!t]
    \centering
    \begin{subfigure}{0.4\textwidth}
        \centering
        \includegraphics[width=\linewidth]{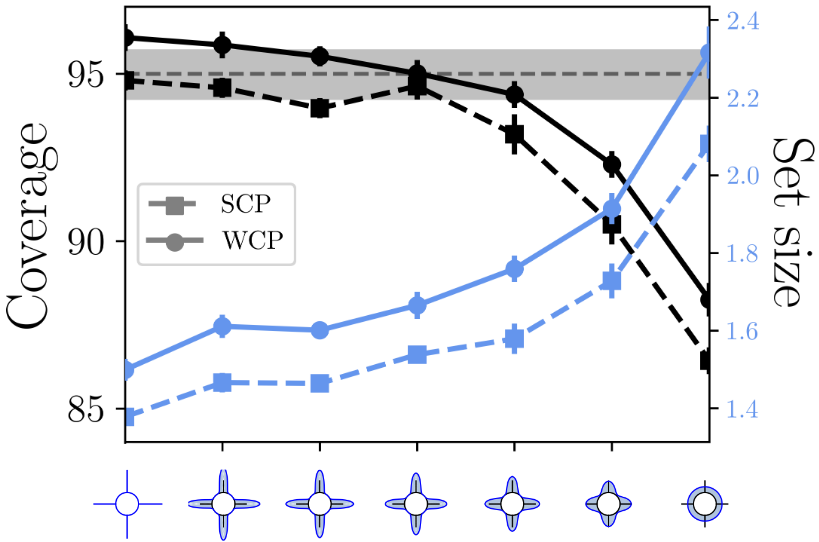}
        \caption{$C4$ to $SO(2)$.}
        \label{fig:subfig1}
    \end{subfigure}
    % \hfill
    \qquad \quad
    \begin{subfigure}{0.4\textwidth}
        \centering
        \includegraphics[width=\linewidth]{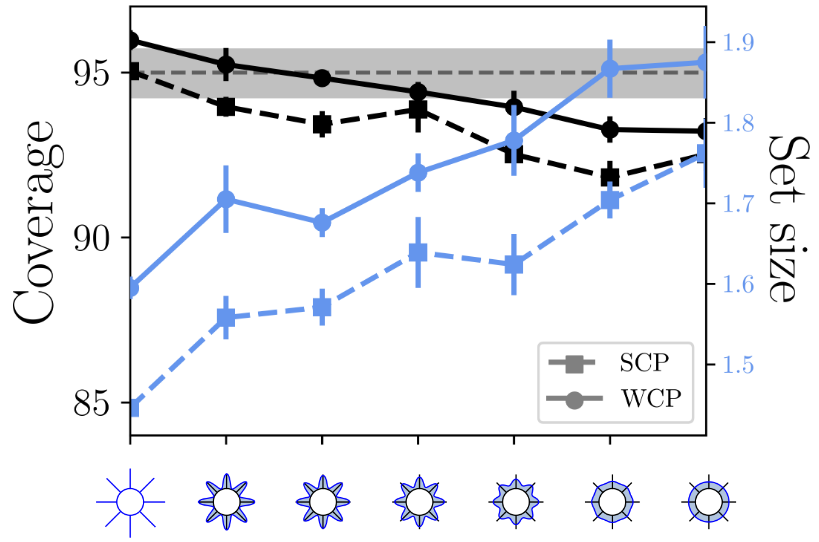}
        \caption{$C8$ to $SO(2)$.}
        \label{fig:subfig2}
    \end{subfigure}
    \caption{
    Coverage and set sizes for the double shift setting (\autoref{tab:wcp-settings}, third row) on CIFAR-10, for split (SCP) and geometrically weighted conformal prediction (WCP). We train the canonicalization network on $\gD_{cal}$ with $C4$ resp. $C8$ group knowledge, and gradually induce a secondary test shift by interpolating from benign to strong shift on $SO(2)$ (left to right end of the $x$-axis showing group distribution). Eventual coverage breakdown is inevitable, but geometric weighting can delay the effect. 
    }
    \label{fig:weighting}
\end{figure*}

%% file: text/related_work.tex
\paragraph{Model-agnostic equivariance and canonicalization.} Group equivariance in deep learning models is typically realized through architectures that inherently incorporate equivariance constraints, and as such a wide array of equivariant layers exist \citep{cohen2016group, weiler2019general, Bekkers2020B-Spline, cohen2019gauge, finzi2020generalizingconvolutionalneuralnetworks, finzi2021arbitrary, ruhe2023cliffordgroupequivariantneural}. These models generally require a full training procedure and carefully tailored architecture design. In contrast, a novel range of approaches enable model-agnostic equivariance, integrating it into pretrained backbones with minimal training or finetuning. These methods generally fall into three categories. First, symmetrization methods apply group averaging operators to non-equivariant base models to enforce equivariance, as explored by \citep{basu2023equituninggroupequivariantfinetuning, basu2023efficientequivarianttransferlearning, kim2024learningprobabilisticsymmetrizationarchitecture}. Second, frame-averaging techniques, such as those discussed in \citep{puny2021frameaveraging, duval2023faenetframeaveragingequivariant, atzmon2021frameaveraging}, focus on identifying efficient yet expressive subsets of groups for averaging. The third category encompasses canonicalization methods \citep{kaba23equivariance, mondal2023equivariant, panigrahi2024improved}, which offer a competitive and resource-efficient alternative to the first two. Unlike symmetrization and frame-averaging, canonicalization utilizes an auxiliary network to provide explicit per-sample group estimates.

\paragraph{Conformal prediction under distribution shift.} A substantial body of recent work has explored the handling of non-exchangeable data sequences, including time series by tracking and adapting miscoverage rates \citep{Gibbs2021AdaptiveCI, angelopoulos2024online, zaffran2022adaptive, angelopoulos2024conformalpid, prinzhorn2024conformal} or employing different weighting strategies \citep{barber2023conformal, guan2023localized, amoukou2023adaptive}. Efforts to directly address shift settings have considered covariate shift \citep{tibshirani2019conformal}, label shift \citep{podkopaev2021distribution}, and broader generalizations \citep{prinster2024conformal}, with various likelihood ratio-based weights. However, to the best of our knowledge no specific handling of \emph{geometric} shifts has been explored in the literature. The only work explicitly utilizing geometry for conformal prediction is \citet{kaur2022idecode}, who employ a notion of equivariance to detect out-of-distribution samples.

%% file: text/discussion.tex
We propose leveraging geometric information to supplement conformal prediction, robustifying the procedure against \emph{geometric} data shifts and ensuring fundamental conditions such as exchangeability are preserved. We explore multiple applications on integrating the approaches: mitigating performance drops due to geometric variations (\autoref{subsec:exp-robust}), employing it as a diagnostics tool for conditional coverage (\autoref{subsec:exp-condcover}), and as a weighting mechanism in settings involving multiple shifts (\autoref{subsec:exp-weightcp}). While we instantiate our approach using the canonicalization principle, the underlying methodology is broadly applicable and should extend to any sample-wise geometry extractor providing similar group information.

\paragraph{Limitations and Outlook.} Our work predominantly explores shifts caused by rotation groups, following \cite{mondal2023equivariant, kaba23equivariance} and related works. While theoretically extendable to other groups like roto-reflections, practical implementation across broader groups remains unexplored. Indeed, other research has highlighted the challenges with continuous canonicalization \citep{dym2024equivariantframesimpossibilitycontinuous}, aligning with practical difficulties observed by \citep{mondal2023equivariant} in its application within the image domain. Alternatively, CP could be combined with symmetrization or frame averaging techniques, of which variants exists that use group weighting mechanisms \citep{dym2024equivariantframesimpossibilitycontinuous, kim2024learningprobabilisticsymmetrizationarchitecture}. Such weighting could potentially be used as an alternative to extract per-sample group estimates.

In \autoref{subsec:exp-weightcp}, generalization to unseen group elements assumes probability mass is concentrated on nearby elements, providing relevant information. While often valid, per-sample inaccuracies in the canonicalization network's 
group predictions (see \eg~\autoref{tab:cifar10-miscanonicalization}) can still negatively impact conformal results. More generally, the inclusion of a \emph{trained} geometry extractor as an additional `bolt-on' module means the pipeline is susceptible to errors propagating down-stream, but can also greatly benefit from future improvements in that regard. Future work can also explore the use of geometric information for more conformal prediction settings including regression tasks \citep{sesia2021conformal}, stream data \citep{Gibbs2021AdaptiveCI} or more general risk notions \citep{angelopoulos2024crc}. A more in-depth exploration of WCP could help develop robuster geometric weighting schemes, perhaps through \cite{prinster2024conformal}'s lens as weighted permutations. Ultimately, the intersection of conformal prediction and geometric deep learning remains largely unexplored, offering promising directions for future work.

%% file: text/ack_impact.tex
\begin{contributions} 
    PvdL and AT contributed jointly to ideation and methodology, and co-authored the paper. In addition, PvdL initiated the preliminary approach and conducted all experiments, while AT initiated the theoretical motivation. EB contributed to ideation development, project guidance and feedback.
\end{contributions}

\begin{acknowledgements} 
    We thank Siba Smarak Panigrahi for valuable discussions about the canonicalization codebase upon which this work relies. We also thank Rajeev Verma and Sharvaree Vadgama for insightful discussions on geometric uncertainty.
\end{acknowledgements}

%% file: text/appendix.tex
\onecolumn

\begin{center}
    {\Large\bfseries CP$^2$: Leveraging Geometry for Conformal Prediction via Canonicalization}\\[0.5em]
    {\Large\bfseries --- Supplementary Material ---}
\end{center}

\label{appendix}

\tableofcontents

\newpage

\section{Additional Experiment Details}
\label{app:details}

\paragraph{Considerations on the canonicalization network.} It is important to stress the motivation of a light-weight, \emph{post-hoc} method when making architectural design choices about the canonicalization network (CN). As such we desire an efficient and thus smaller and potentially less expressive CN. Naturally, a small performance drop over the symmetry-unaware predictor is expected by incorporating the CN, since as a trained model it is prone to (some) pose prediction errors itself. Additionally, both the limited expressivity of the light-weight CN as well as the resolution on $SO(2)$ induces discretization artifacts. We examine the CN's miscanonicalization in \autoref{tab:cifar10-miscanonicalization}, reporting the fraction of correctly predicted group elements on CIFAR-10 data subject to $C4$ and $C8$ rotations. We additionally nuance that the miscanonicalization results do not directly translate to down-stream lower prediction accuracy, as the predictor itself can exhibit robustness to minor pose variations (\eg~correctly classifying images with rotation angles in $[-45^\circ, 45^\circ]$) and hence may still correctly predict a target despite erroneous pose alignment. 

\begin{table}[h]
  \caption{
  Fraction of group elements correctly predicted by trained canonicalization networks on CIFAR-10. We subsequently evaluate the models on a hold-out split of data under the same group effects, \ie~$C4$ or $C8$.
  }
  \centering
  \begin{tabular}{l|c}
    \toprule
     \textbf{Model} & \textbf{\% Corr. angles} \\
     \midrule
    Canonicalization with $(G=4)$ & 87.46\\
    Canonicalization with $(G=8)$ & 87.23\\     
    \bottomrule
  \end{tabular}
  \label{tab:cifar10-miscanonicalization}
\end{table}

\subsection{Robustness to Geometric Data Shifts}
\label{app:details-robust}

\paragraph{Image canonicalization network.} For the image domain, when taking into account the aforementioned desire for an efficient geometric module, we restrict ourselves to $C4$ and $C8$-equivariant canonicalization networks. We adopt the models described in \cite{mondal2023equivariant}, employing a compact 3-layer, $G$-equivariant WideResNet, where $G$-equivariance is achieved through the use of E2CNN \citep{weiler2019general}. All models are trained for a maximum of 100 epochs with early stopping, and optimized using Adam.

\paragraph{Point cloud canonicalization network.} For the continuous point cloud domain, we similarly follow the approaches outlined in \cite{mondal2023equivariant} by adopting a compact Vector Neuron model \citep{deng2021vnn}. These models are trained for 250 epochs with a cosine learning rate scheduler, and optimized using Adam.

\subsection{Diagnostics for Conditional Coverage}
\label{app:details-mcp}

\paragraph{Intuition.} We explore the hypothesis that different group elements (\eg~particular rotation angles) can correlate with specific data partitions due to their distinct geometric properties \citep{urbano2024selfsupervised, vanderlinden2025learningsymmetriesweightsharingdoubly, allingham2024generativemodelsymmetrytransformations, romero2023learningpartialequivariancesdata}. For example, isotropic shapes such as a ring or the digit “0” (\eg~in MNIST) may withstand arbitrary rotations without altering their class identity or losing significant visual features. Hence their geometric pose may be naturally uniformly distributed over the rotation group. Conversely, shapes such as the digit “6” transform into a “9” when rotated at 180$^\circ$, potentially leading to erroneous prediction. Consequently, one would not expect to observe such group elements to meaningfully contribute to the shape's natural pose distribution. Our experiments in \autoref{subsec:exp-condcover} manually induce such shifts (\eg~on class labels) in order to highlight the canonicalization network's accurate recovery of such geometric behaviour.

\paragraph{Experimental design.} We induce several group shifts conditioned on particular target partitions:
\begin{itemize}
    \item \texttt{dirac}: A dirac distribution over the group, pinpointing a single group element per partition;
    \item \texttt{normal}: A normal distribution over the group; and
    \item \texttt{var-gauss}: various Gaussian distributions with standard deviations in $[0.0001, 0.001, 0.01, 0.1, 1.0, 10.0]$.
\end{itemize}

To improve the visual recovery of partition-conditional group effects, we additionally exclude data points for which the canonicalization network's predicted group probability falls below a predefined threshold, ensuring that only samples with confident group predictions are taken into account. This aids in counteracting some of the canonicalization's erroneous predictions (see \autoref{tab:cifar10-miscanonicalization}) to better demonstrate why mondrian conformal prediction may be useful when clear patterns exist.

\subsection{Weighting for Double Shift Settings.}
\label{app:details-wcp}

\paragraph{Intuition.} The canonicalization network’s role is to mitigate the first shift between $\gD_{train}$ and $\gD_{cal}$, but in the double-shift setting of \autoref{subsec:exp-weightcp} we also encounter a subsequent shift between $\gD_{cal}$ and $\gD_{test}$ (see \autoref{tab:wcp-settings}, third row), such as from known discrete group elements in $C8$ to potentially any rotation in  $SO(2)$ (see \autoref{fig:weighting}, right). From empirical observations and prior studies, minor rotations (e.g., within $\pm 5$ degrees) have shown to enhance the accuracy of down-stream pose prediction tasks. This improvement is often attributed to the alignment with natural object variations captured in datasets \citep{mondal2023equivariant}. This insight suggests that within small deviations from known group elements, a well-trained CN is capable of accurately identifying the nearest group element. This accuracy decreases as the continuous rotation deviates further from these discretized elements, reaching maximum ambiguity at positions equidistant from two neighboring group elements (\ie~maximal shift). Therefore, when a test sample’s transformation is close to one of these discretized rotations, the CN tends to assign higher probabilities to that group element or its immediate neighbors. We harness these insights on the CN’s probabilistic output to obtain geometry-informed weights for weighted conformal prediction, enhancing robustness to rotations not explicitly covered by the discrete, known group elements.

\paragraph{Experimental design.} To navigate the transition between the (exchangeable) discrete setting and the (non-exchangeable) uniform $SO(2)$ group, we model a group distribution on the sphere. Specifically, we define it as either discrete peaks at the $C4$ or $C8$ elements, or as a continuous distribution using a mixture of `von Mises' distributions each centered at $C4$ or $C8$ elements. The `von Mises' p.d.f. is of the form $f(x \,|\, \mu, \kappa) = 1/(2\,\pi\, I_0(\kappa)) \cdot\mathrm{exp}(\kappa \mathrm{ cos}(x))$, where $\mu$ denotes a location parameter and $\kappa$ controls the concentration of mass around $\mu$. Varying $\kappa$ facilitates the interpolation between discrete $C4$ or $C8$ sampling and more uniform $SO(2)$ sampling. In \autoref{fig:weighting}, we use $\kappa=[50, 40, 30, 20, 10]$ as interpolative factors, and visualize the resulting spherical group distributions on the $x$-axis. Regarding the inverse geometric weighting relationship described in \autoref{subsec:method-usecases}, we visualize different values of the modulating parameter $p$ and their effect on the weighting distribution in \autoref{fig:weights_pow}; and their impact on the $C4$ to $SO(2)$ double-shift setting in \autoref{fig:weights_params}. For the main paper, we opt for a cross-entropy distance metric and set $p=2.0$ as it empirically displays a good trade-off between set size and coverage target.

\input{fig/fig_weight_parameters}

\section{Additional Experiment Results}
\label{app:exp}

\input{fig/fig_proxy_partition_cov}

\input{fig/fig_mcp_cov_both}

\paragraph{Robustness to geometric shift with Thr \citep{sadinle2019least}.} The following tables display accuracy and conformal results for the geometric shift experiments from \autoref{subsec:exp-robust} on both images and point clouds using another conformal scoring approach (Thr \citep{sadinle2019least}). This conformal scoring function is simply defined as $s(\vx_i, \vy_i) = 1 - \hat{p}(\rvy_i = \vy_i | \vx_i)$ for any true class label $\vy_i$. Results are consistent with those in the main paper using \emph{APS} \citep{romano2020classificationvalidadaptivecoverage}.

\input{tab/robust_shift_pointcloud_thr}
\input{tab/robust_shift_cifar10_aps}
\input{tab/robust_shift_cifar10_thr}

\input{tab/robust_shift_cifar100_thr}

%% file: fig/fig_weight_parameters.tex
\begin{figure}[H]
    \centering
    \includegraphics[width=0.4\linewidth]{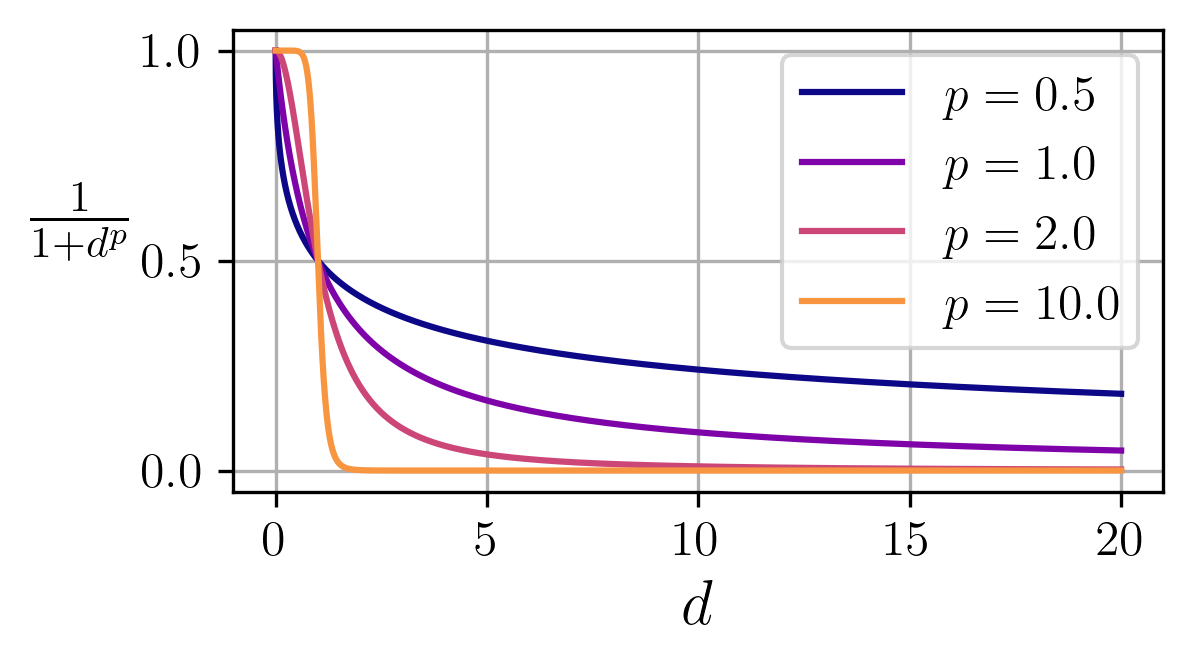}
    \caption{Illustration of the impact of the modulation parameter $p$ on the inverse geometric weighting relation. As $p$ increases, the weighting distribution approaches a binary mask for smal distances.}
    \label{fig:weights_pow}
\end{figure}

\begin{figure}[H]
    \centering
        \centering
        \includegraphics[width=0.6\linewidth]{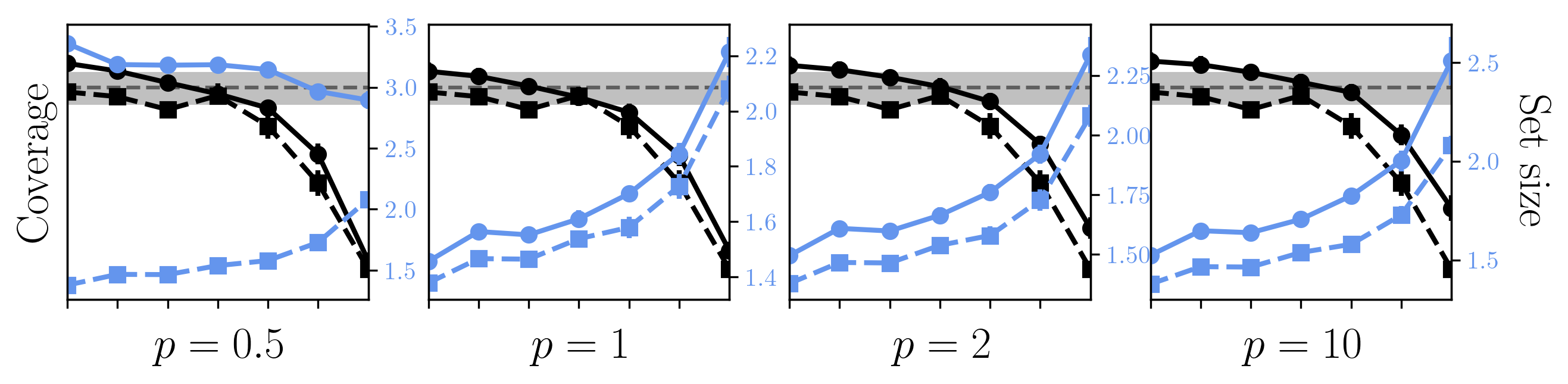}
        \includegraphics[width=0.6\linewidth]{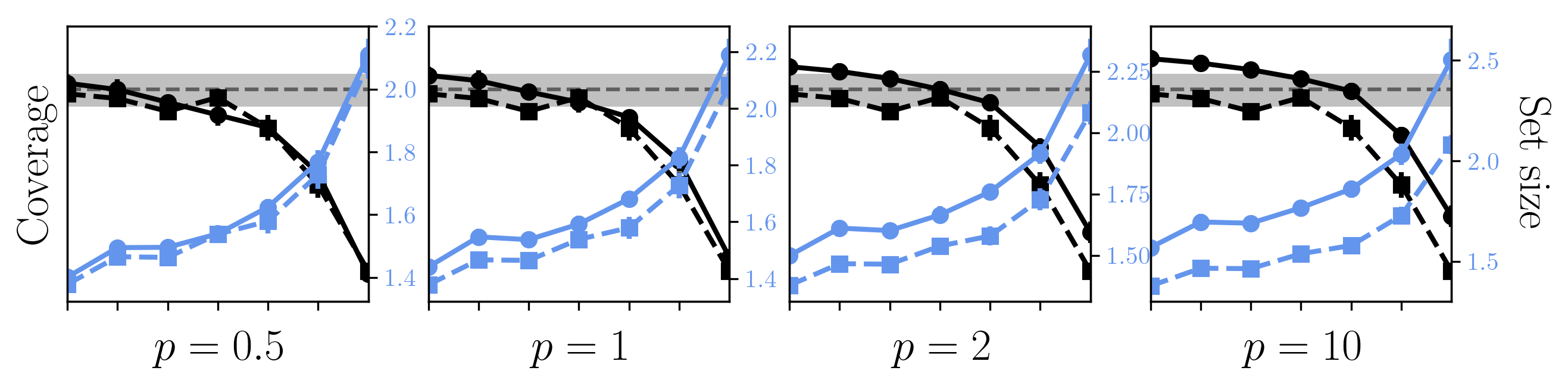}
    \caption{Ablation on the modulation parameter $p$ and two different distributional distance metrics (\emph{top}: KL-divergence, \emph{bottom}: cross-entropy) for the double-shift setting (\autoref{subsec:exp-weightcp}) for $C4$ to $SO(2)$.}
    \label{fig:weights_params}
\end{figure}

%% file: fig/fig_proxy_partition_cov.tex
\begin{figure}[H]
    \centering
    % First subfigure
    \begin{subfigure}{0.33\textwidth}
        \centering
        \includegraphics[width=\linewidth]{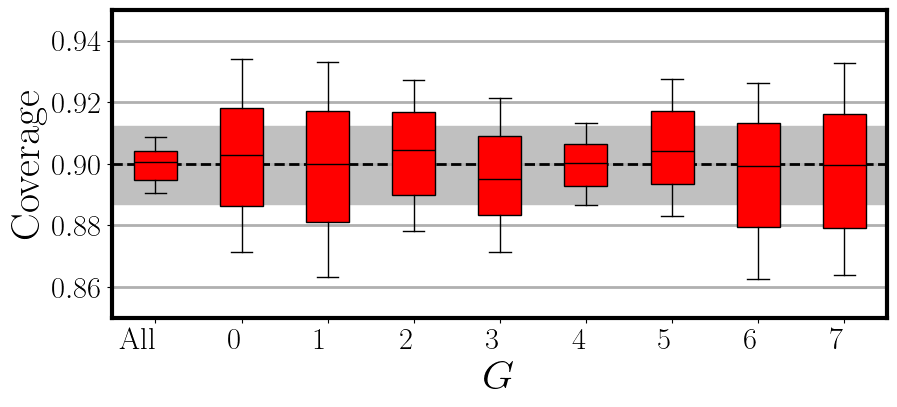}
        % \caption{Weighting mechanism 1.}
        % \label{fig:subfig1}
    \end{subfigure}
    % \hfill
    % Second subfigure
    \begin{subfigure}{0.33\textwidth}
        \centering
        \includegraphics[width=\linewidth]{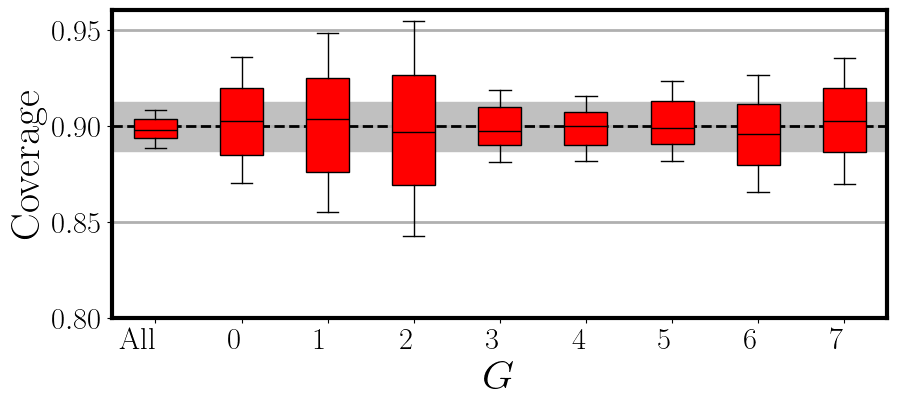}
        % \caption{Weighting mechanism 2.}
        % \label{fig:subfig2}
    \end{subfigure}
    \begin{subfigure}{0.33\textwidth}
        \centering
        \includegraphics[width=\linewidth]{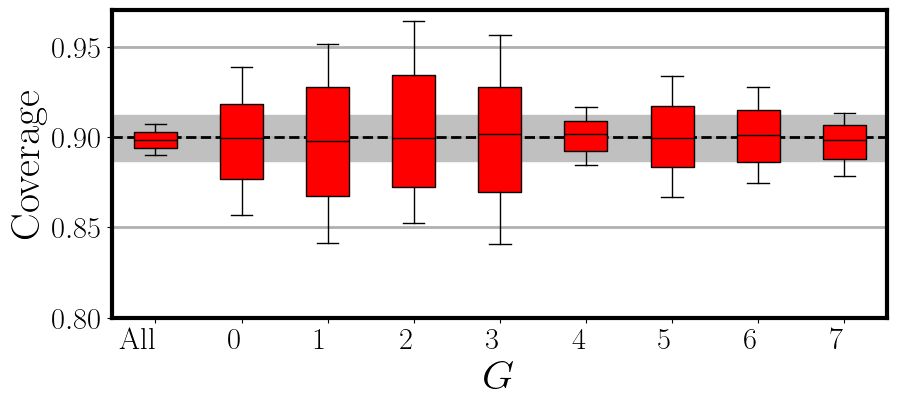}
        % \caption{Weighting mechanism 2.}
        % \label{fig:subfig2}
    \end{subfigure}
    
    \caption{
     Empirical coverage for mondrian conformal prediction on the exact geometric group partitions for shift settings 2, 3 and 4 in \autoref{fig:class_conditional_mondrian}. As we perform MCP directly on these partitions, per-group target coverage is nominally guaranteed.
    }
    \label{fig:app-partition_coverage}
\end{figure}

%% file: fig/fig_mcp_cov_both.tex
\begin{figure*}[!h]
    \centering
    % First subfigure
    \begin{subfigure}{0.32\textwidth}
        \centering
        \includegraphics[width=\linewidth]{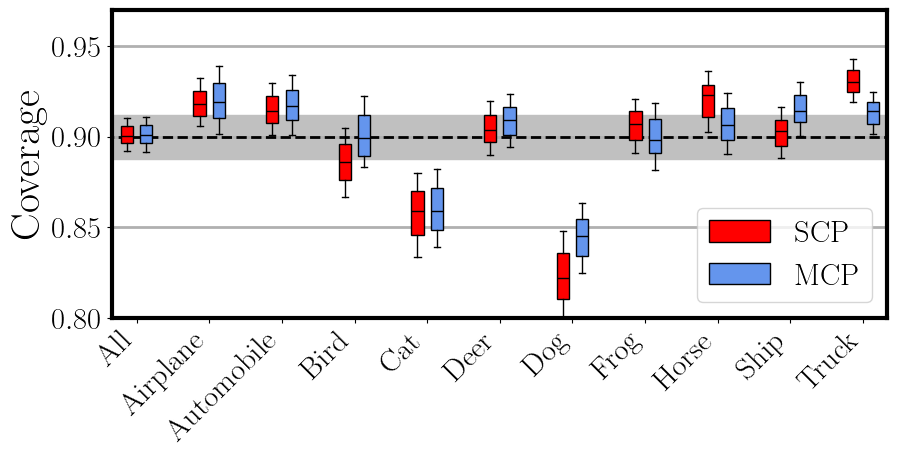}
        % \caption{Weighting mechanism 1.}
        \label{fig:subfig1}
    \end{subfigure}
    % Second subfigure
    \begin{subfigure}{0.33\textwidth}
        \centering
        \includegraphics[width=\linewidth]{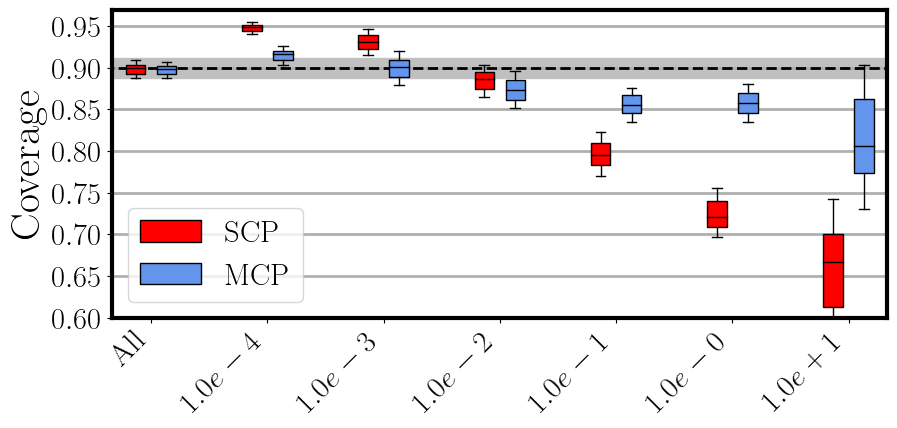}
        % \caption{Weighting mechanism 2.}
        \label{fig:subfig2}
    \end{subfigure}
    \begin{subfigure}{0.33\textwidth}
        \centering
        \includegraphics[width=\linewidth]{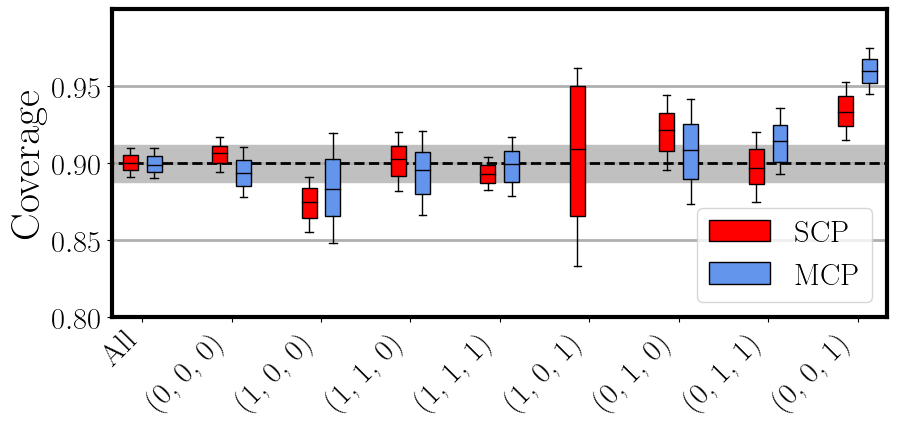}
        % \caption{Weighting mechanism 2.}
        \label{fig:subfig2}
    \end{subfigure}
    \caption{
    Empirical coverage for the target partitions via split (SCP) and mondrian conformal prediction (MCP) for shift settings 1, 3 and 4 in \autoref{fig:class_conditional_mondrian}. Improved coverage balance is obtained \emph{by proxy} when a strong and accurately caputed geometric relationship is apparent. \emph{Left}: SCP and MCP match when all target partitions exhibit identical group distributions. \emph{Center, right}: MCP yields more balanced coverage in the presence of a meaningful proxy relation.
    }
    \label{fig:target_partition_coverage}
\end{figure*}

%% file: tab/robust_shift_pointcloud_thr.tex
\begin{table}[H]
  \caption{
  Results with \textbf{Thr \citep{sadinle2019least} on ModelNet-40} for target coverage $(1-\alpha) = 95\%$ across $SO(3)$ rotation shift. Target coverage is efficiently maintained when no shift occurs, but grows excessively large and uninformative (\colorbox{gray!25}{\textcolor{gray!25}{oo}}) when the underlying predictor $\hat{f}_{\vtheta}$ is not equivariant. Results are reported across $T=10$ random calibration and test splits. \# Param. indicates the required number of trained parameters. (*) Use slightly different train/test splits and data preprocessing.
  }
  \centerline{
  \setlength{\tabcolsep}{7pt}
  \resizebox{\linewidth}{!}{
      \begin{tabular}{ll|ccc|ccc}
        \toprule
         & & \multicolumn{3}{c}{\textbf{No Shift}} & \multicolumn{3}{c}{\textbf{$SO(3)$ Rotation Shift}}  \\
         \textbf{Model} & \# Param. & Acc & Cov & Set Size & Acc & Cov & Set Size \\
         \midrule
         PointNet & 0.7 M
         & 87.49 & $ 95.11 \pm .006 $ & $ 1.368 \pm .019 $ 
         & 8.73  & $ 100.00 \pm .000 $ & \cellcolor{gray!25}$ 40.000 \pm .000 $ \\
         DGCNN & 1.8 M
         & 91.41 & $ 94.68 \pm .005 $ & $ 1.103 \pm .015 $ 
         & 15.15 & $ 100.00 \pm .000 $ & \cellcolor{gray!25}$ 40.000 \pm .000 $ \\
         % 2
         Rapidash* & 1.7 M
         & 86.59 & $ 95.12 \pm .010 $ & $ 1.498 \pm .078 $ 
         & 12.12 & $ 100.00 \pm .000 $ & \cellcolor{gray!25}$ 40.000  \pm .000 $ \\
         \midrule
         PointNet + $SO(3)$ & 0.7 M
         & 57.46 & $ 95.3 \pm .008 $ & $ 6.532 \pm .384 $ 
         & 55.63 & $ 94.85 \pm .011 $ & $ 7.058 \pm .497 $ \\
         DGCNN + $SO(3)$ &  1.8 M
         & 86.55 & $ 95.18 \pm .004 $ & $ 1.508 \pm .040 $ 
         & 85.74 & $ 94.71 \pm .010 $ & $ 1.554 \pm .105 $ \\
         % 3
         Rapidash* + $SO(3)$ & 1.7 M
         & 76.50 & $ 95.07 \pm .005 $ & $ 3.519 \pm .212 $ 
         & 75.93 & $ 95.01 \pm .007 $ & $ 3.323 \pm .208 $ \\
         \midrule
         % 1
         Invariant Rapidash* + $SO(3)$  & 1.7 M
         & 74.47 & $ 95.00 \pm .008 $ & $ 4.525 \pm .368 $ 
         & 74.35 & $ 95.21 \pm .009 $ & $ 4.650 \pm .306 $ \\
         
         % 4
         Equivariant Rapidash* + $SO(3)$ &  2.0 M
         & 88.21 & $ 95.05 \pm .006 $ & $ 1.446 \pm .038 $ 
         & 87.60 & $ 95.14 \pm .006 $ & $ 1.416 \pm .032 $ \\
         \midrule
         PRLC-PointNet & 1.3 K
         & 62.11 & $ 95.60 \pm .010 $ & $ 5.936 \pm .361 $ 
         & 62.07 & $ 95.09 \pm .008 $ & $ 5.745 \pm .266 $ \\
         PRLC-DGCNN & 1.3 K
         & 85.78 & $ 95.42 \pm .005 $ & $ 1.663 \pm .052 $ 
         & 85.86 & $ 94.55 \pm .012 $ & $ 1.594 \pm .083 $ \\
        \bottomrule
      \end{tabular}
      }
  }
  \label{tab:pointcloud-robust-thr}
\end{table}

%% file: tab/robust_shift_cifar10_aps.tex
\begin{table}[H]
  \caption{
  Results with \textbf{APS \citep{romano2020classificationvalidadaptivecoverage} on CIFAR-10} for target coverage $(1-\alpha) = 95\%$ across different rotation shifts. Target coverage is efficiently maintained when no shift occurs, but grows excessively large and uninformative (\colorbox{gray!25}{\textcolor{gray!25}{oo}}) when the underlying predictor $f_{\vtheta}$ is not equivariant, or if the wrong group is specified (controlling for $C4$ but exposed to $C8$). Results are reported across $T=10$ random calibration and test splits. \# Param. indicates the required number of \emph{trained} parameters.
  }
  \centerline{
  \setlength{\tabcolsep}{7pt}
  \resizebox{\linewidth}{!}{
      \begin{tabular}{ll|ccc|ccc|ccc}
        \toprule
         & & \multicolumn{3}{c}{\textbf{No Shift}} & \multicolumn{3}{c}{\textbf{$C4$ Rotation Shift}}  & \multicolumn{3}{c}{\textbf{$C8$ Rotation Shift}} \\
         \textbf{Model} & \# Param. & Acc & Cov & Set Size & Acc & Cov & Set Size & Acc & Cov & Set Size \\
         \midrule
         $\hat{f}_{\theta}$ & 23.5 M
         & 92.82 & $ 94.94 \pm .004 $ & $ 1.137 \pm .005 $ 
         & 52.50 & $ 95.06 \pm .005 $ & \cellcolor{gray!25}$ 5.324 \pm .078 $ 
         & 45.39 & $ 95.28 \pm .005 $ & \cellcolor{gray!25}$ 6.420 \pm .154 $ 
         \\
         $\hat{f}_{\theta}$ + $SO(2)$ Aug. & 23.5 M
         & 84.97 & $ 94.82 \pm .004 $ & $ 1.560 \pm .019 $ 
         & 85.03 & $ 95.01 \pm .004 $ & $ 1.534 \pm .024 $ 
         & 83.89 & $ 94.99 \pm .006 $ & $ 1.541 \pm .027  $ \\
         $\hat{f}_{\theta}$ + $C_4$ Aug. & 23.5 M
         & 88.08 & $ 95.07 \pm .005 $ & $ 1.362 \pm .019 $ 
         & 87.81 & $ 95.28 \pm .003 $ & $ 1.356 \pm .022 $ 
         & 70.56 & $ 95.11 \pm .004 $ & \cellcolor{gray!25}$ 4.213 \pm .059 $ \\
         $\hat{f}_{\theta}$ + $C_8$ Aug. & 23.5 M
         & 86.39 & $ 95.20 \pm .003 $ & $ 1.477 \pm .018 $ 
         & 86.41 & $ 95.11 \pm .005 $ & $ 1.447 \pm .022 $
         & 85.36 & $ 95.11 \pm .005 $ & $ 1.459 \pm .025 $ \\
         \midrule
         CP$^2$ with $G$=4   & 0.25 M  
         & 88.09 & $ 94.74 \pm .005 $ & $ 1.374 \pm .017 $ 
         & 88.09 & $ 95.10 \pm .004 $ & $ 1.387 \pm .017 $ 
         & 66.80 & $ 94.76 \pm .005 $ & $ 3.814 \pm .087 $ \\
         CP$^2$ with $G$=8  & 0.51 M
         & 88.13 & $ 94.93 \pm .003 $ & $ 1.369 \pm .015 $ 
         & 88.13 & $ 95.02 \pm .003 $ & $ 1.374 \pm .015 $ 
         & 86.97 & $ 94.95 \pm .003 $ & $ 1.462 \pm .020 $ \\
        \bottomrule
      \end{tabular}
      }
  }
  \label{tab:cifar10-robust}
\end{table}

%% file: tab/robust_shift_cifar10_thr.tex
\begin{table}[H]
  \caption{
    Results with \textbf{Thr \citep{sadinle2019least} on CIFAR-10} for target coverage $(1-\alpha) = 95\%$ across different rotation shifts. Target coverage is efficiently maintained when no shift occurs, but grows excessively large and uninformative (\colorbox{gray!25}{\textcolor{gray!25}{oo}}) when the underlying predictor $\hat{f}_{\vtheta}$ is not equivariant, or if the wrong group is specified (controlling for $C4$ but exposed to $C8$). Results are reported across $T=10$ random calibration and test splits. \# Param. indicates the required number of trained parameters.
  }
  \centerline{
  \setlength{\tabcolsep}{7pt}
  \resizebox{\linewidth}{!}{
      \begin{tabular}{ll|ccc|ccc|ccc}
        \toprule
         & & \multicolumn{3}{c}{\textbf{No Shift}} & \multicolumn{3}{c}{\textbf{$C4$ Rotation Shift}}  & \multicolumn{3}{c}{\textbf{$C8$ Rotation Shift}} \\
         \textbf{Model} & \# Param. & Acc & Cov & Set Size & Acc & Cov & Set Size & Acc & Cov & Set Size \\
         \midrule
         $\hat{f}_{\theta}$ & 23.5 M
         & 92.82 & $ 94.79 \pm .004 $ & $ 1.061 \pm .008 $ 
         & 52.50 & $ 95.28 \pm .003 $ & \cellcolor{gray!25}$ 5.171 \pm .050 $ 
         & 45.39 & $ 94.50 \pm .006 $ & \cellcolor{gray!25}$ 6.187 \pm .114 $ 
         \\
         $\hat{f}_{\theta}$ + $SO(2)$ Aug. & 23.5 M
         & 85.00 & $ 94.73 \pm .004  $ & $ 1.485 \pm .022 $ 
         & 85.03 & $ 95.05 \pm .004 $ & $ 1.490 \pm .021 $ 
         & 83.90 & $  95.01 \pm .004  $ & $ 1.476 \pm .020  $ \\
         $\hat{f}_{\theta}$ + $C_4$ Aug. & 23.5 M
         & 88.08 & $  94.92 \pm .005 $ & $ 1.304 \pm .024 $ 
         & 87.81 & $ 95.27 \pm .004 $ & $ 1.308 \pm .018 $ 
         & 70.56 & $ 95.11 \pm .004 $ & \cellcolor{gray!25}$ 4.077 \pm .055 $ \\
         $\hat{f}_{\theta}$ + $C_8$ Aug. & 23.5 M
         & 86.39 & $ 95.14 \pm .002 $ & $ 1.430 \pm .012 $ 
         & 86.41 & $ 94.97 \pm .005 $ & $ 1.402 \pm .025 $
         & 85.36 & $ 95.08 \pm .004  $ & $ 1.417 \pm .020  $ \\
         \midrule
         CP$^2$ with $G$=4   & 0.25 M  
         & 88.12 & $ 94.75 \pm .003 $ & $ 1.340 \pm .012  $ 
         & 88.12 & $ 95.03 \pm .003 $ & $ 1.351 \pm .018 $ 
         & 66.84 & $ 94.90 \pm .005 $ & $ 3.742 \pm .067 $ \\
         CP$^2$ with $G$=8  & 0.51 M
         & 88.27 & $ 94.90 \pm .003 $ & $ 1.329 \pm .017 $ 
         & 88.27 & $ 94.92\pm .002 $ & $ 1.331 \pm .015 $ 
         & 87.04 & $ 94.78 \pm .003 $ & $ 1.417 \pm .017 $ \\
        \bottomrule
      \end{tabular}
      }
  }
  \label{tab:cifar10-robust-thr}
\end{table}

%% file: tab/robust_shift_cifar100_thr.tex
\begin{table}[H]
  \caption{
    Results with \textbf{Thr \citep{sadinle2019least} on CIFAR-100} for target coverage $(1-\alpha) = 95\%$ across different rotation shifts. Target coverage is efficiently maintained when no shift occurs, but grows excessively large and uninformative (\colorbox{gray!25}{\textcolor{gray!25}{oo}}) when the underlying predictor $\hat{f}_{\vtheta}$ is not equivariant, or if the wrong group is specified (controlling for $C4$ but exposed to $C8$). Results are reported across $T=10$ random calibration and test splits. \# Param. indicates the required number of trained parameters.
  }
  \centerline{
  \setlength{\tabcolsep}{7pt}
  \resizebox{\linewidth}{!}{
      \begin{tabular}{ll|ccc|ccc|ccc}
        \toprule
         & & \multicolumn{3}{c}{\textbf{No Shift}} & \multicolumn{3}{c}{\textbf{$C4$ Rotation Shift}}  & \multicolumn{3}{c}{\textbf{$C8$ Rotation Shift}} \\
         \textbf{Model} & \# Param. & Acc & Cov & Set Size & Acc & Cov & Set Size & Acc & Cov & Set Size \\
         \midrule
         $\hat{f}_{\theta}$ & 23.7 M
         & 71.66 & $ 95.15 \pm .003 $ & $ 5.57 \pm .102 $ 
         & 40.17 & $ 100.00 \pm .000 $ & \cellcolor{gray!25}$ 100.000 \pm .000 $ 
         & 33.68 & $ 100.00 \pm .000 $ & \cellcolor{gray!25}$ 100.000 \pm .000 $ 
         \\
         
         $\hat{f}_{\theta}$ + $SO(2)$ Aug. & 23.7 M
         & 60.13 & $ 95.00 \pm .005 $ & $ 10.614 \pm .407 $ 
         & 59.72 & $  95.06 \pm .005 $ & $ 10.895 \pm .383 $ 
         & 58.27 & $ 95.05 \pm .005 $ & $ 10.534 \pm .433 $ \\
         
         $\hat{f}_{\theta}$ + $C_4$ Aug. & 23.7 M
         & 63.03 & $ 95.08 \pm .006 $ & $ 9.361 \pm .428 $ 
         & 62.72 & $ 95.16 \pm .004 $ & $ 9.399 \pm .328 $ 
         & 49.72 & $ 100.00 \pm .000 $ & \cellcolor{gray!25}$ 100.000 \pm .000 $ \\
         $\hat{f}_{\theta}$ + $C_8$ Aug. & 23.7 M
         & 62.53 & $ 95.35 \pm .005 $ & $ 10.063 \pm .424 $ 
         & 62.37 & $ 95.27 \pm .003  $ & $ 9.990 \pm .180 $ 
         & 60.82 & $ 95.26 \pm .003  $ & $ 9.856 \pm .171 $ \\
         \midrule
         CP$^2$ with $G$=4  & 1.0 M
         & 65.48 & $ 95.20 \pm .005 $ & $ 10.173  \pm .491 $ 
         & 65.48 & $ 95.02 \pm .004 $ & $ 9.982 \pm .388 $
         & 48.33 & $ 98.46 \pm .024 $ & \cellcolor{gray!25}$ 78.834 \pm 32.332 $ \\
         
         CP$^2$ with $G$=8   & 2.0 M
         & 65.51 & $ 95.39 \pm .005 $ & $ 10.674 \pm .465 $ 
         & 65.51 & $ 94.99 \pm .004 $ & $ 10.271 \pm .401 $ 
         & 64.16 & $ 94.96 \pm .004 $ & $  11.10 \pm .312 $ \\
        \bottomrule
      \end{tabular}
      }
  }
  \label{tab:cifar100-robust-thr}
\end{table}